\documentclass[conference]{IEEEtran}
\def\BibTeX{{\rm B\kern-.05em{\sc i\kern-.025em b}\kern-.08em
    T\kern-.1667em\lower.7ex\hbox{E}\kern-.125emX}}

\usepackage{cite}
\usepackage{amsmath,amssymb,amsfonts}
\usepackage{graphicx}
\usepackage{textcomp}
\usepackage[table,xcdraw]{xcolor}
\usepackage{url}
\usepackage{hyperref}
\usepackage{amsmath}
\usepackage{algpseudocode}
\usepackage{algorithm}
\usepackage{caption}
\usepackage{subcaption}
\usepackage{multirow}
\usepackage{booktabs}
\usepackage{lipsum}
\usepackage{comment}
\usepackage{rotating}
\usepackage{multicol}

\usepackage[normalem]{ulem} 

\usepackage{enumitem}

\begin{document}

\title{D3FL: Data Distribution and Detrending for Robust Federated Learning in Non-linear Time-series Data
}
\author{
    \IEEEauthorblockN{Harsha Varun Marisetty}
    \IEEEauthorblockA{\textit{Department of CSIS} \\
    \textit{BITS-Pilani, Hyderabad} \\
    p20200437@hyderabad.bits-pilani.ac.in}
    \and
    \IEEEauthorblockN{Manik Gupta}
    \IEEEauthorblockA{\textit{Department of CSIS} \\
    \textit{BITS-Pilani, Hyderabad} \\
    manik@hyderabad.bits-pilani.ac.in}
    \and
    \IEEEauthorblockN{Yogesh Simmhan}
    \IEEEauthorblockA{\textit{Department of CDS} \\
    \textit{Indian Institute of Science, Bengaluru} \\
    simmhan@iisc.ac.in}
}

\maketitle

\begin{abstract}
 With advancements in computing and communication technologies, the Internet of Things (IoT) has seen significant growth. IoT devices typically collect data from various sensors, such as temperature, humidity, and energy meters. Much of this data is temporal in nature. Traditionally, data from IoT devices is centralized for analysis, but this approach introduces delays and increased communication costs. Federated learning (FL) has emerged as an effective alternative, allowing for model training across distributed devices without the need to centralize data.

In many applications, such as smart home energy and environmental monitoring, the data collected by IoT devices across different locations can exhibit significant variation in trends and seasonal patterns. Accurately forecasting such non-stationary, non-linear time-series data is crucial for applications like energy consumption estimation and weather forecasting. However, these data variations can severely impact prediction accuracy.
The key contributions of this paper are: (1) Investigating how non-linear, non-stationary time-series data distributions, like generalized extreme value (gen-extreme) and log norm distributions, affect FL performance. (2) Analyzing how different detrending techniques for non-linear time-series data influence the forecasting model's performance in a FL setup.

We generated several synthetic time-series datasets using non-linear data distributions and trained an LSTM-based forecasting model using both centralized and FL approaches. Additionally, we evaluated the impact of detrending on real-world datasets with non-linear time-series data distributions. Our experimental results show that: (1) FL performs worse than centralized approaches when dealing with non-linear data distributions. (2) The use of appropriate detrending techniques improves FL performance, reducing loss across different data distributions.
  
  \end{abstract}






\section{Introduction}
\label{sec:intro}
With the improvements in computing and communication technology, there is an increase in the number of data producers, such as Internet of Things (IoT) devices. Statista estimates that there will be 32.1 billion IoT devices in 2030 \cite{statista}. With the increased availability of IoT devices, there has been an increase in applications like surveillance \cite{zhang2024yolo}, traffic monitoring \cite{dui2024iot}, pollution sensing \cite{ramadan2024real}, energy usage monitoring \cite{anusha2024internet}, etc.

A significant portion of the data collected from sensors, such as those measuring humidity, temperature, gas levels, and energy consumption, is temporal in nature. Time-series forecasting plays a crucial role in various applications, such as estimating the energy load of buildings \cite{taik2020electrical}, predicting vehicle traffic volumes for intelligent traffic systems \cite{xie2023machine}, and forecasting pollution levels \cite{zhang2023meta}. Accurate time-series forecasting is essential for informed decision-making in areas like energy management and urban traffic control. For instance, precise energy load predictions enable better grid management, optimizing energy distribution. Similarly, accurate traffic forecasting aids in improving congestion management strategies. The data generated by IoT devices can vary widely and exhibit distinct time-series characteristics, including trends and seasonality. In traditional time-series analysis, trends and seasonality are often removed to make the data stationary, which is better suited for statistical modeling \cite{pawar2022techniques}.

Traditionally, to develop a forecasting model with a network of multiple IoT devices generating data, all the data from the devices would be centralized. However, this approach leads to increased communication costs, delays, and potential privacy concerns regarding users' data. To address these issues, Google introduced Federated Learning (FL) \cite{federated,McMahan2016CommunicationEfficientLO}, where model parameters are trained locally on each device, and only the local models are sent to the server, rather than raw data.

Time-series data generated by multiple devices, such as building electricity consumption or vehicular traffic, is often user-dependent and follows a non-iid (non-independent and identically distributed) pattern \cite{fekri2022distributed}. FL is generally less effective when dealing with non-iid data \cite{wu2021fast}. While LSTMs have been widely used for time-series forecasting in FL settings \cite{subramanya2021centralized}, \cite{huang2023new}, \cite{perifanis2023federated}, \cite{de2024forecasting}, most studies overlook the impact of non-iid behavior caused by specific time-series characteristics, such as trends, on the performance of FL.

We identify three key gaps in the existing literature on time-series forecasting using Federated Learning (FL):
(1) Although several studies have explored FL based time-series forecasting models~\cite{wen2022solar,skianis2023data,perry2021energy,fekri2022distributed,dasari2021privacy,de2024forecasting,qu2023personalized,makonin2019hue,zhang2022federated}, they largely overlook the impact of non-linear data distributions at individual clients, both in FL and centralized setups.
(2) While LSTM-based models are commonly used in time-series forecasting, prior work typically applies preprocessing techniques like detrending and seasonality adjustment to ensure data stationarity~\cite{hyndman2008forecasting,box2015time,Zhang2005}. Detrending, for example, has proven effective in load forecasting using ARIMA models~\cite{vuksanovic2017load}. However, it remains unclear whether such preprocessing is beneficial or even appropriate in FL contexts.
(3) Among the available detrending techniques suited for non-linear, non-stationary time-series data, no prior study systematically evaluates which method is most effective when training LSTM-based forecasting models within an FL framework.


To address these gaps, we investigate the impact of different data distributions with non-linear characteristics across clients, apply various detrending techniques, and train an LSTM model using Federated Learning (FL). We assess the performance of the resulting model by evaluating the forecasting results on each client's local dataset.

Our main contributions are as follows:
\begin{enumerate}[leftmargin=*]
\item We emphasize the importance of non-linear data distributions on time-series forecasting models trained using FL.
\item We observe a significant decline in FL performance when the data is not detrended, leading to much higher loss values. In comparison, a centralized training approach outperforms FL in this regard.
\item We assess different detrending techniques and their impact on the forecasting model trained with FL. Our findings show that the effectiveness of detrending varies across different non-linear data distributions. Therefore, the choice of technique should be based on the underlying characteristics of the data. We evaluate the impact of detrending on both synthetic and real-world datasets.
\end{enumerate}

The rest of the paper is organized as follows. In Section~\ref{sec:background}, we review FL, time-series characteristics and various detrending techniques. Next, in Section~\ref{sec:related}, we discuss the existing works on forecasting models trained using FL. Then, in Section~\ref{sec:methodology}, we describe the experimental setup, synthetic data generation, real world dataset and the LSTM based model trained for forecasting. In Section~\ref{sec:results}, we discuss the performance of the forecasting model with different data distributions and the impact of detrending techniques on the forecasting model. Lastly, we conclude with our key insights in Section~\ref{sec:conclusion}.

\vspace{-5pt}


\section{Background}
\label{sec:background}
In this section, we first describe FL and later elaborate on time-series characteristics.

\begin{figure}[t]
    \centering
    \includegraphics[width=0.8\linewidth]{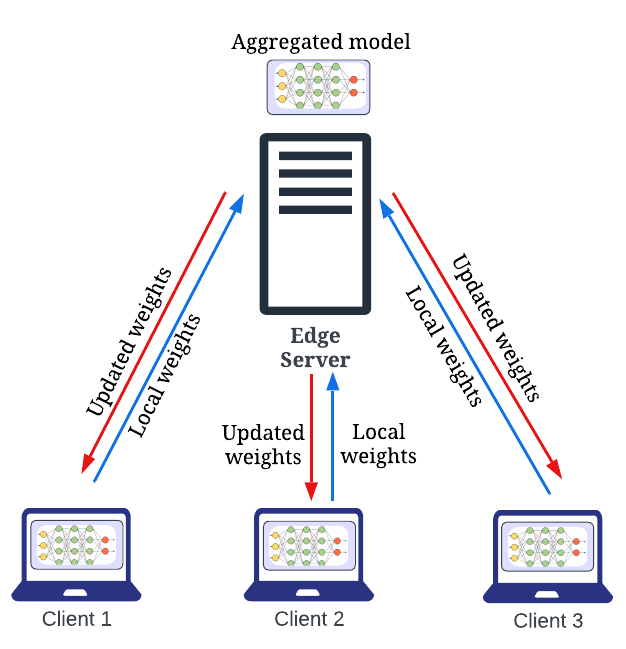}
    \caption{Federated learning system design}
    \label{fig:FL}
\end{figure}

\subsection{Federated Learning (FL)}
FL is a collaborative machine learning technique where clients aim to train a single model without centrally sharing their data. To achieve this, the participating clients train a model that has been initialized by the server on their local data, over one or more epochs, and return the updated \textit{local model} weights to the server
(Figure~\ref{fig:FL}). A simple weighted average over the model weights at the server gives the aggregated \textit{global model}, termed as \textit{FedAvg}~\cite{federated}. This updated global model is sent back to the clients for another \textit{round} of local training, and this process continues till the configured number of training rounds or global model convergence. Compared to a centralized approach, where the data and the model training are centralized, the FL approach reduces the use of communication resources for transferring large local datasets and improves the privacy of the user's local data, while also using distributed computing resources at the clients.


    

\subsection{Time-Series Characteristics}
\begin{figure}[t]
    \centering
    \includegraphics[width=0.99\linewidth]{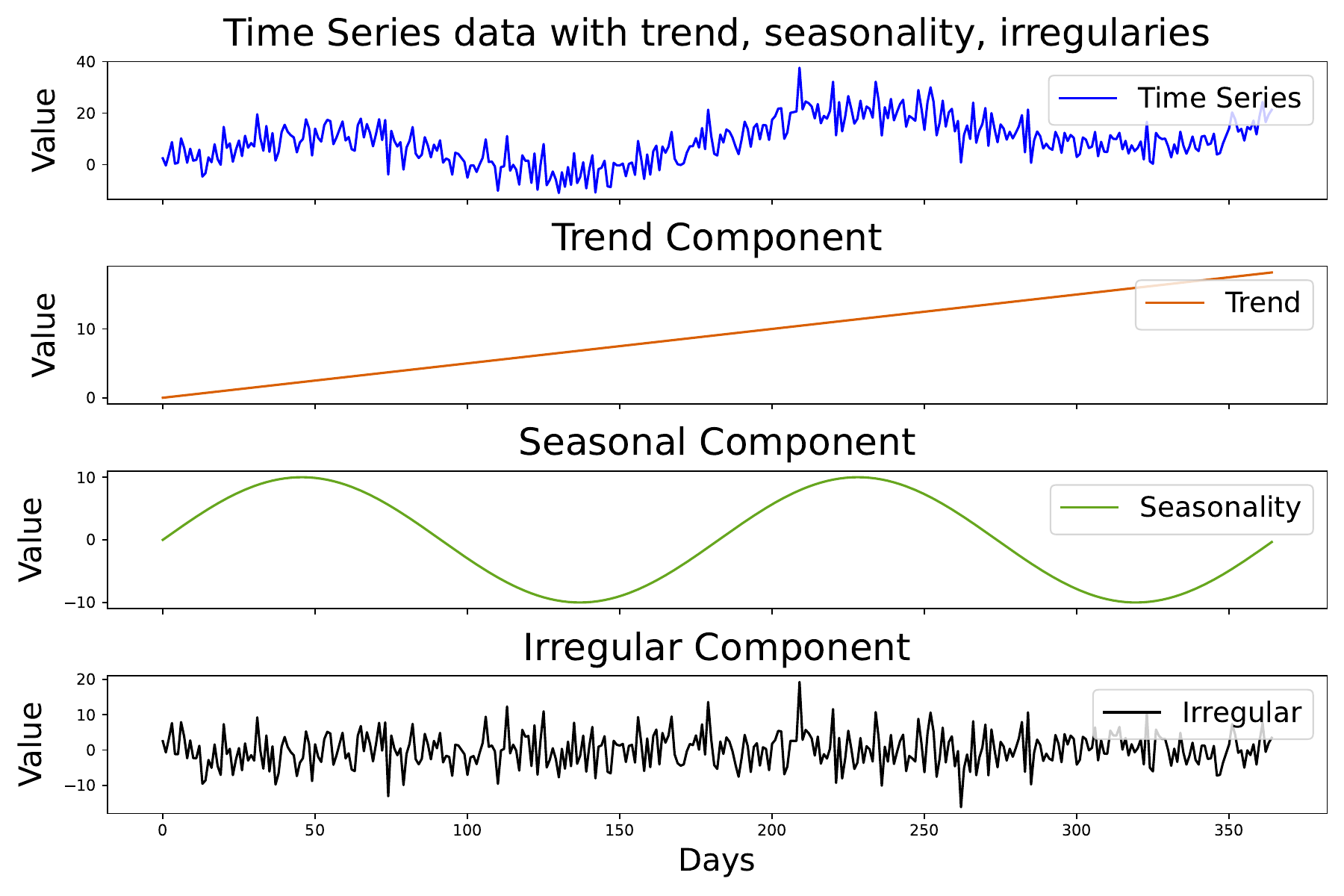}
    \caption{Various time-series data characteristics}
    \label{fig:tscomp}
    \vspace{-10pt}
\end{figure}

Time-series data is a sequence of data points where each data point $x_t$ is associated with a time t and is represented as $
\{x_t : t \in T\} $ where $T$ is a set of time indices and $x$ is the variable that changes over time. Time-series data can be classified as either \textit{stationary} or \textit{non-stationary}. The mean, variance, and auto covariance in stationary data do not change over time, but they can change in non-stationary data. The characteristics of non-stationary time-series data are as follows~\cite{hansen2003forecasting}: 
\begin{itemize}[leftmargin=*]
    \item \textbf{Trend} - Trend indicates how the data is varying over time, i.e., whether it is increasing or decreasing, or if there is no trend (stationary). For example, increasing retail store sales for three years indicates an upward trend.
    \item \textbf{Seasonality} - Seasonality refers to the repeating patterns in data. For example, increase in retail store sales annually every holiday season. 
    \item \textbf{Irregularity} - Irregular and random variations in data that can not be attributed to either trends or seasonality. For example, the announcement of a ``Special Sale'' at a retail store that increases sales.
\end{itemize}
    Figure \ref{fig:tscomp} visualizes a time-series data (top) and its characteristics such as trend, seasonality and irregularities (bottom three).

Further, trends can be classified either as \textit{linear} or \textit{non-linear} trends. In linear trends, the data change rate is consistent, like a line. Meanwhile, in non-linear trends, the rate of change in data is inconsistent, like a curve. A non-linear trend may follow a quadratic or polynomial curve. 
In general, it has been observed that air pollution and climate data like extreme weather events, temperatures, etc., as well as financial data such as stock prices, exhibit non-linear trends. In particular, it has been shown that extreme weather data and temperatures follow a \textbf{gen-extreme distribution}~\cite{rypkema2021modeling}. In contrast, financial data mostly follow either \textbf{log norm or log gamma distribution}~\cite{werner_hrlimann_2001,jimichi2018visualization}.

Next, we present details of the commonly occurring data distributions for non-linear time-series data, usually exhibited by climate and financial applications:
    \subsubsection{Generalized
    Extreme Value (gen-extreme) Distribution}
    Gen-extreme value distribution is a type of probability distribution used for modelling extreme values (maximum or minimum) over a particular time interval. It is used to model rare events like floods, droughts, financial losses, etc.
    Three parameters characterize the distribution:
    \begin{itemize}
        \item \textit{Location parameter (\(\mu\)):} Shifts the distribution along the x-axis.
        \item \textit{Scale parameter (\(\sigma\)):} Stretches or compresses the distribution.
        \item \textit{Shape parameter (\(\xi\)):} Tunes the tail of the distribution.
    \end{itemize}
    When \(\xi = 0\), it follows \textit{Gumbel} distribution; \(\xi > 0\) corresponds to the \textit{Fréchet} distribution; and \(\xi < 0\) corresponds to the \textit{Weibull} distribution. 
    The probability density function is defined by Equation \ref{eq:gev}: 
    \begin{equation}
        \begin{split}
        A = \frac{1}{\sigma} \left(1 + \xi \frac{x - \mu}{\sigma}\right)^{-\left(\frac{1}{\xi} + 1\right)} \\
        f(x; \mu, \sigma, \xi) = A \cdot  \exp \left\{-\left(1 + \xi \frac{x - \mu}{\sigma}\right)^{-\frac{1}{\xi}}\right\}
        \end{split}
        \label{eq:gev}
    \end{equation}

   \subsubsection{Log Normal Distribution}
   A log norm distribution is a probability distribution of a variable whose logarithm is normally distributed. 
   Positive numbers such as profits, sales of retail shops, etc., fall under the log norm distribution, and the distribution is right-skewed.
   The log norm distribution has two parameters:
   \begin{itemize}
       \item \(\mu\): The mean of the natural logarithm of the variable.
        \item \(\sigma\): The standard deviation of the natural logarithm of the variable.
   \end{itemize}
   The log norm probability density function is defined as per Equation \ref{eq:lognorm}:
   \begin{equation}
   f(x; \mu, \sigma) = \frac{1}{x \sigma \sqrt{2 \pi}} \exp \left\{-\frac{(\ln x - \mu)^2}{2 \sigma^2}\right\}
    \label{eq:lognorm}
   \end{equation}
   

\subsection{Detrending Techniques}

Detrending is an important preprocessing step to remove the trends and make the data stationary. 
Various detrending techniques can be applied to time-series data before evaluating the forecasting model performance. Say we have a time-series data with \(T\) instances, \( x = \{x_0, x_1, x_2, x_3, \ldots, x_T\} \), where \( X_i \) is the value of the time-series at time \( i \). A few detrending techniques are described below for which the forecasting model performance has been compared.

    \subsubsection{Differencing}
    Differencing involves subtracting the current value from the previous value. The differenced value at time t \(X_t'\) is given by Equation \ref{eq:diff}. \begin{equation}X_i' = X_i - X_{i-1} \label{eq:diff}\end{equation} 
    
    \subsubsection{Moving Average}
    As the name suggests, this technique makes use of a sliding window with a sliding window period \( p \leq i \) to compute the average of the sliding window and remove it from the original series, as shown in Equation \ref{eq:mavg}. 
    \begin{equation}
    X_i' = X_i - \frac{1}{n} \sum_{j=0}^{p-1} X_{i-j}
    \label{eq:mavg}
    \end{equation}
    
    \subsubsection{Detrend by Removal Mean}
    In this method, the mean of the data \(\overline{X}\) is removed from every instance of time-series data, as shown in Equation \ref{eq:mean}.
    \begin{equation} 
    \begin{split}
        \overline{X} = \frac{1}{n} \sum_{t=1}^{n} X_t \\
        X'_i = X_i - \overline{X}
    \end{split}
    \label{eq:mean}
    \end{equation}
    
    \subsubsection{Detrend with a Linear Model}
    This method involves fitting a linear regression model \(L_i\) over the time-series data and then subtracting the fitted values to remove the trend from the data \( X'_i \), as given in Equation \ref{eq:linear}.
    \begin{equation} 
    \begin{split}
        L_i = \beta_0 + \beta_1 i + \epsilon_i \\
        X'_i = X_i - L_i 
    \end{split}
    \label{eq:linear}
    \end{equation}
    
    \subsubsection{Detrend with a Quadratic Model}
    This method involves fitting a quadratic regression model \(Q_i\) over the time-series data and then subtracting the fitted values to remove the trend from the data \( X'_i \), as listed in Equation \ref{eq:quad}.
    \begin{equation} 
    \begin{split}
        Q_i = \beta_0 + \beta_1 i + \beta_2 i^2 + \epsilon_i \\
        X'_i = X_i - Q_i 
    \end{split}
    \label{eq:quad}
    \end{equation}


\section{Related Work}\label{sec:related}
In this section, we discuss the significant works on time-series modelling techniques for forecasting and recent works that use FL to train the forecasting model for time-series data.


\subsection{Time-Series Modelling}
The traditional statistical approaches to model time-series data are based on exponential smoothing \cite{hyndman2008forecasting} and autoregressive integrated moving average (ARIMA) \cite{box2015time}. However, they perform better on stationary data. Feed-forward neural network-based models for forecasting are unable to capture the seasonality and trend of the data \cite{Zhang2005}, but perform better when the trend and seasonality is removed. The later works continued the removal of trends and seasonality even for other models such as RNN, LSTM and transformer based models such as Informer \cite{zhou2021informer}, which could capture long term dependencies. But they failed to investigate if these models could capture the non-linear trends and seasonality well or not. We address this gap in an FL setting.

\subsection{Federated Learning-based Time-series Forecasting}
Traditionally, the data from multiple sources is centralized, and the forecasting model training is done. Considering the privacy restrictions on the source data and bandwidth required to centrally accumulate them, FL was proposed as an alternative for training a distributed model. A few recent works on time-series forecasting models trained using FL are summarized below.

Several works involve detrending as part of their data preparation before FL. Wen at al.~\cite{wen2022solar} train a spatial and temporal attention-based neural network using FL approach for solar forecasting, however the seasonality has been removed. For energy load forecasting, others~\cite{taik2020electrical} train an LSTM model by removing the seasonality in Pecan Street Inc.'s Dataport household data and has also considered the weather data of the locality for better forecasting.
Fedtime~\cite{skianis2023data} is proposed for building a load forecasting model. The authors have used differencing to remove the trend, and during aggregation, Fedtime penalized the local models using aged data. So that the local models trained with recent time-series data would contribute more to the global model, the authors could achieve lesser loss with this method.


There are also several FL timeseries forecasting papers that do not consider detrending of the dataset.
Perry et al.~\cite{perry2021energy} prepare a forecasting model for forecasting the electric vehicle (EV) charging station network load. The forecasting model was LSTM-based, trained using the FL approach. The EV stations are separated geographically, so they clustered the EV stations with the help of the dynamic time warping technique applied to usage patterns. They perform cluster-based aggregation. However, the trend was not removed from the forecasting model.
Others~\cite{fekri2022distributed} have performed electricity load prediction by training an LSTM based forecasting model based on FL approach. The forecasting model was prepared using London Hydro, a local electrical distribution utility. They normalize the data using min-max normalization; however, the trend was not removed. 
Some have trained a deep neural network (DNN) based forecasting model on ASHRAE dataset for energy load prediction following FL approach~\cite{dasari2021privacy}. They could achieve less root mean square log error, However, detrending was not part of the preprocessing.

Datasets from Smart* and Building Data Genome Project 2 have been used to train a CNN-LSTM based forecasting model using a federated learning approach~\cite{de2024forecasting}. The CNN layer captured the hidden features and passed them to the LSTM layer. They augmented Smart* with the help of Generative Adversarial Networks (GANs). But detrending was not applied to original data nor to the augmented data.
Similarly, others~\cite{qu2023personalized} have trained a LSTM based model on HUE dataset~\cite{makonin2019hue} with a personalized FL approach that reduced the prediction error on client's data. Some have trained CNN-Attention-LSTM based model for the campus electricity usage~\cite{zhang2022federated}.
Sievers and Blank~\cite{SieversTransformerVsLSTMFL2023} have compared the performance of CNN, LSTM and transformer in a FL setup and the transformer could achieve lesser loss than the rest. However, the trend was not removed.

In summary, based on our literature review, we find that some of the FL works apply detrending; however, most of the existing FL works do not remove trends or seasonality from the time-series data. Hence, the need for detrending time-series data is unclear and has not been examined properly for FL setup. Further, the role of non-linear data distributions has also not been considered. As a part of this paper, we aim to research and fill the gap by examining how detrended, non-linear time-series data distributions affect the performance of forecasting models trained using a federated learning approach.

\section{Proposed Methodology}
\label{sec:methodology}
This section discusses our FL setup, synthetic dataset creation, description of real-world datasets used and the model details.

\subsection{FL Setup}

We follow a centralized FL approach with a single server aggregating updates from multiple clients. Our experimental setup includes two machines: a workstation with an Intel i9-12900K processor with 24 vCPU cores and 64GB of memory, and a desktop with an Intel i7-10700 processor with 16 vCPU cores and 16GB of memory. The workstation and desktop are directly connected via a single network hop, operating under ideal conditions with no jitter or delay.
The Flotilla framework~\cite{flotilla} has been used for running FL on edge devices and Docker containers. We have used Docker containers to simulate the FL server and the client machines. The server container is hosted on the workstation and configured with 4GB RAM and two vCPU cores. The clients are hosted on the workstation and desktop, and each is configured with 2GB RAM and one vCPU core. We have spawned ten client containers, five on the workstation and five on the desktop, and one server container for the experiments. This paper primarily investigates model performance for forecasting non-linear time series data, with and without detrending, under both federated and centralized training setups. To isolate the impact of detrending and training paradigms, we assume ideal network conditions and do not consider variations in network environments.

To check the FL performance on time-series data with non-linear trends, we have used vanilla federated learning, which uses \textit{FedAvg} as an aggregator~\cite{federated}. The performance of the global model that is trained using the FL approach is then compared to that of a model trained centrally, with all the data centralized at the server.

In our experiments, the FL model is trained for 100 global rounds, and in each global round, clients are trained for one local epoch based on their data. At each client, the local data has been chronologically sorted and the first 90 percent of the data has been used for training and the remaining 10 percent of data has been used as validation data. Each client trains on the training data and checks the global model performance on their validation data. Mean Square Error (MSE), Mean Absolute Error (MAE), and Root Mean Square Error (RMSE) are used for evaluating the performance of the trained forecasting models.

\begin{figure}[t]
        \centering
        \includegraphics[width=0.8\linewidth]{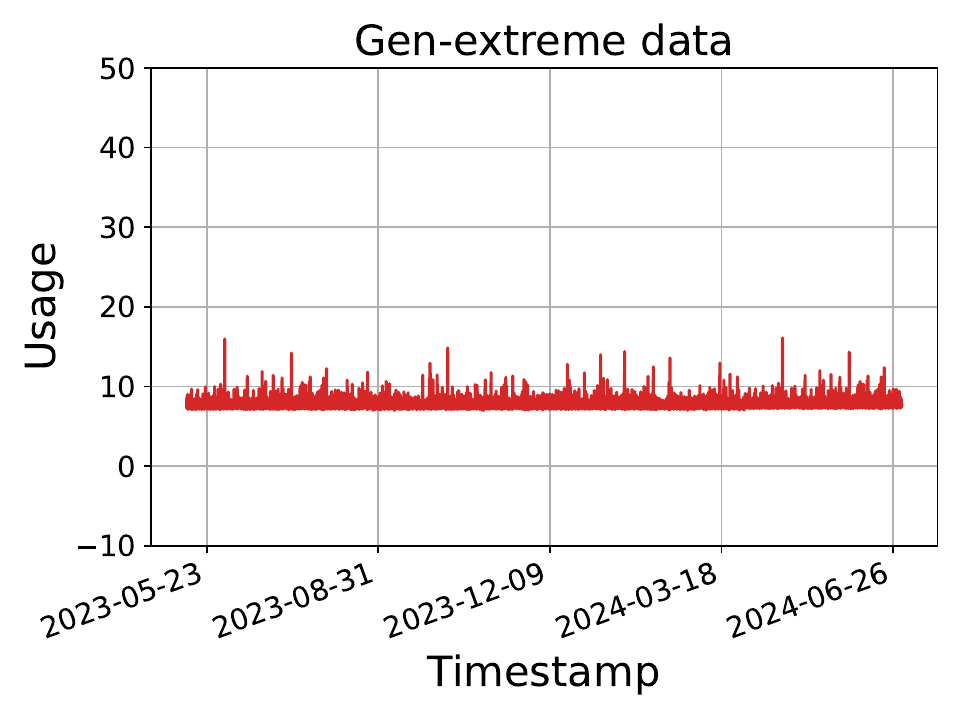}
        \caption{Sample synthetic data for a client with gen-extreme distribution without detrending}
        \label{fig:c1_data_exp_1}
        \vspace{-10pt}
\end{figure}

\begin{figure}[t]
        \centering
        \includegraphics[width=0.8\linewidth]{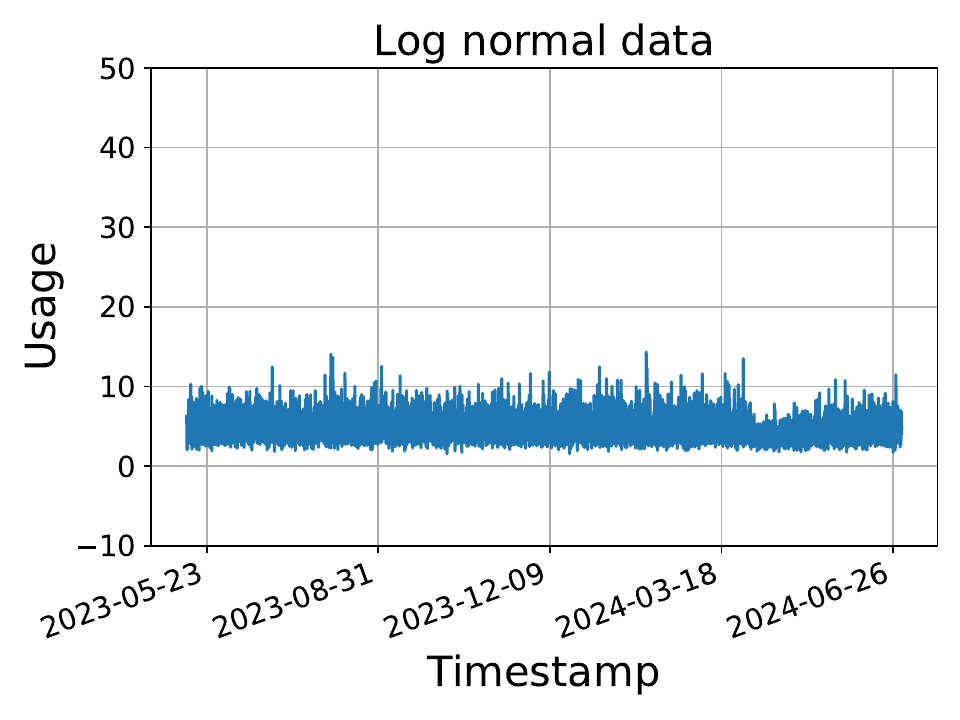}
        \caption{Sample synthetic data for a client with log norm distribution without detrending}
        \label{fig:c8_data_exp_1}
        \vspace{-10pt}
\end{figure}

\subsection{Synthetic Non-linear Time-Series Dataset Generation}
\label{sec:synthetic_data}

To evaluate the FL performance of time-series data with gen-extreme and log norm distributions, 20 univariate datasets are generated using the numpy and scipy packages, specifically, 10 for each type of data distribution. To create a unique dataset for each client, the location or mean is determined using a sine function that takes both the client number and the time step as input. An additional offset is then applied to a portion of the data to introduce non-stationary behaviour. Other parameters such as shape, scale in gen-extreme, and $\mu$ and $\sigma$ of log norm are selected so that the resulting data lies in the range of 2 to 20. Each dataset was created with 10,000 data points, sampled every hour starting from 11-May-2023 9:00 AM till 01-Aug-2024 0:00 PM. Synthetic data for a client that follows gen-extreme distribution is visualised in Figure \ref{fig:c1_data_exp_1}, while synthetic data for a client with log-norm distribution is shown in Figure \ref{fig:c8_data_exp_1}.

\subsection{Real World Dataset Selection}

\begin{figure}[!t]
        \centering
        \includegraphics[width=0.8\linewidth]{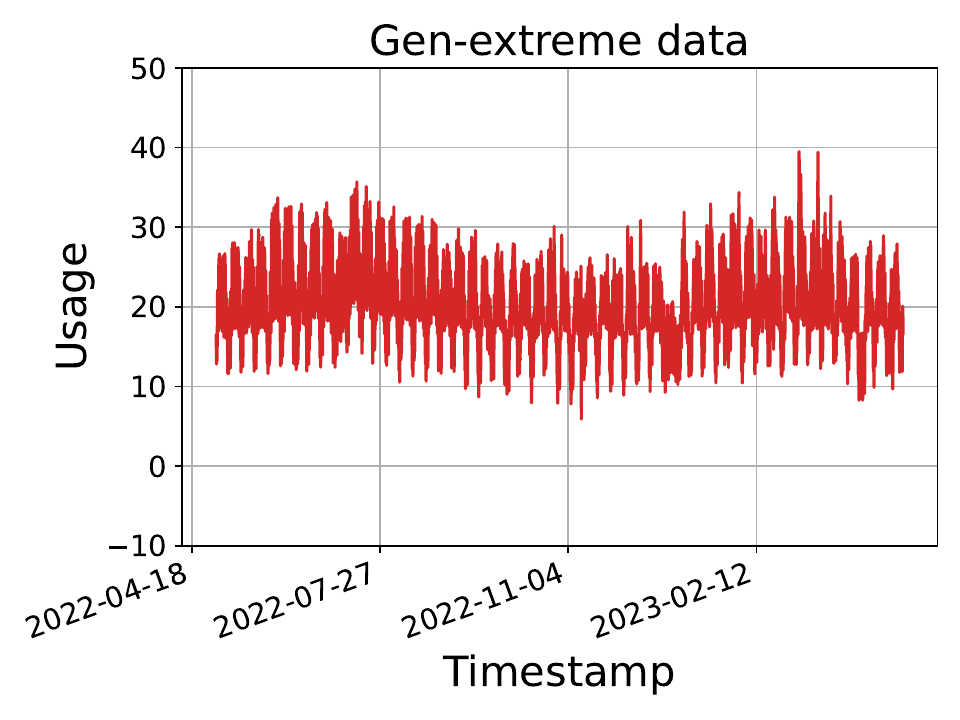}
        \caption{Sample Ausgrid data of a substation with gen-extreme distribution without detrending}
        \label{fig:ausgrid_c1_data_exp_1}
\end{figure}

\begin{figure}[!t]
        \centering
        \includegraphics[width=0.8\linewidth]{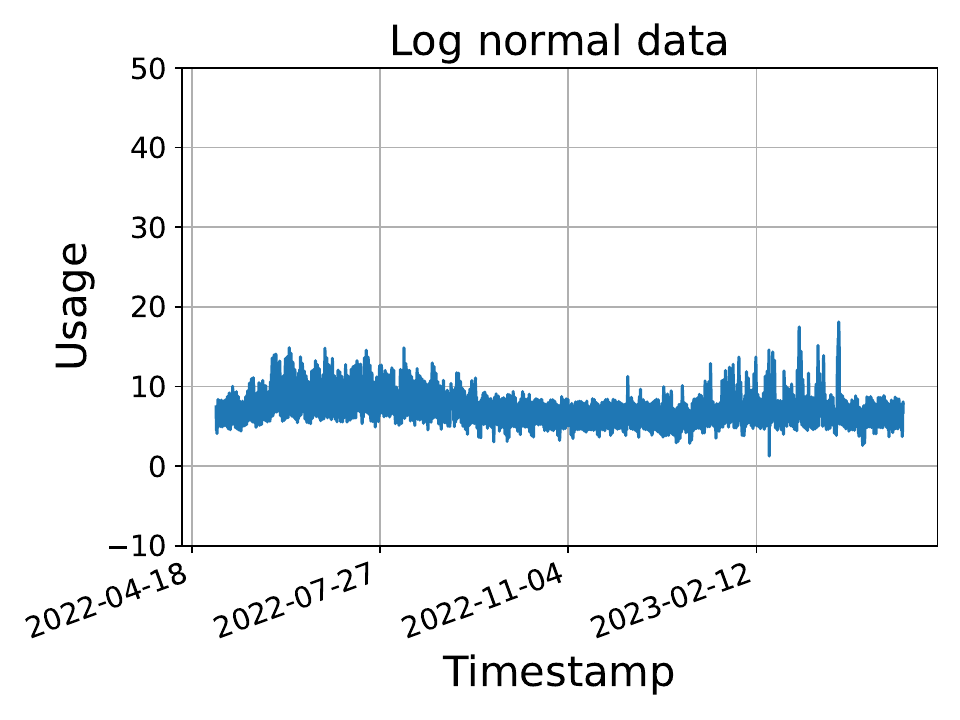}
        \caption{Sample Ausgrid data of a substation with log norm distribution without detrending}
        \label{fig:ausgrid_c8_data_exp_1}
\end{figure}

To evaluate the FL performance with real-world time-series data, we have selected an energy dataset by Ausgrid 2023~\cite{ausgrid}. It has cleaner data with fewer missing values, and 194 substations in Australia to choose clients from, with different distributions. Each substation in has energy usage data recorded from 01-05-2022 00:00 till 2023-04-31 00:00 with a granularity of 15min. The substations' energy usage data have different distributions such as Weibull, \textit{t}, gamma, beta, log gamma, log norm, gen-extreme, etc. To evaluate the FL on log norm and gen-extreme distributions, we selected 10 substations that follow a log norm distribution and 10 substations that follow a gen-extreme distribution. The substations with log norm distribution that are selected are: Avoca, Auburn, Bass Hill, Berowra, Chatswood, Jannali, Mayfield west, Lidcombe, Long Jetty and Olympic park. Similarly, substations selected with gen-extreme distribution are: Balgowlah north, Berkeley vale, Blakehurst, Botany, Charmhaven, Darling harbour, Epping, Jesmond, Milperra and Nulkaba.
Sample data of substations with gen-extreme and log norm are shown in Figure \ref{fig:ausgrid_c1_data_exp_1} and Figure \ref{fig:ausgrid_c8_data_exp_1}, respectively.

\subsection{Time-series Forecasting Using LSTM}
 
While Transformer-based models have revolutionized time series forecasting, they require high computation and memory for self-attention. This makes deploying them on edge devices such as Raspberry Pis and Nvidia Jetson difficult. So among the other popular time-series forecasting models such as gated Recurrent Unit (GRU), Recurrent Neural Network (RNN) and Long Short-Term Memory (LSTM), we have used LSTM in our experiments since it helps overcome the vanishing gradient problem. 

The forecasting model comprises a LSTM layer and an output layer, as shown in Table 
 \ref{tab:lstm_model}. As the data is sampled every hour at the client, our forecasting model forecasts 2h into the future at 1h granularity, based on the past 24h of readings. So, the LSTM layer takes an input sequence of length 24 with 1 feature per time step and produces an output with 128 hidden units. This LSTM layer captures the long-term dependencies within the input sequence. The output from the LSTM layer is fed into a fully connected layer to get a 2-dimensional space from a 128-dimensional space. This corresponds to predicting 2 future time steps based on the input sequence. As the data generated was univariate, the input size was 1. We have used a grid search using the Ray package \cite{ray} for the optimal number of look back and look forwards with the selected parameters as listed in Table \ref{tab:lstm_model}.

\begin{table}[!t]
\caption{LSTM Model Architecture}
\resizebox{\columnwidth}{!}{%
\centering
\begin{tabular}{@{}ccc@{}}
\toprule
\textbf{Layer}                       & \textbf{Parameter}     & \textbf{Details} \\ \midrule
\multirow{2}{*}{\textbf{LSTM Layer}} & Number of Hidden Nodes & 128              \\
                                     & Activation Function    & Tanh             \\ \midrule
\textbf{Fully Connected (FC) Layer}  & Output Size            & 2                \\ \bottomrule
\end{tabular}
}
\label{tab:lstm_model}
\vspace{-10pt}
\end{table}
\vspace{-5pt}
\section{Experimental Details}
The following section outlines the experimental setup used for both synthetic and real-world datasets. Each experiment involves 10 clients and falls into one of the following three data distribution categories:

\begin{itemize}
    \item \textit{Gen-extreme based}: All clients generate data following a generalized extreme value distribution.

\item \textit{log norm based}: All clients use a log norm distribution.

 \item \textit{Mixed (Gen-extreme + Log norm)}: The first 5 clients use a gen-extreme distribution, while the remaining 5 use a log norm distribution.
\end{itemize}

Within each distribution category, we evaluate six preprocessing settings: five detrending techniques—differencing, moving average, mean subtraction, linear detrending, and quadratic detrending—along with a baseline where no detrending is applied. This results in a total of 18 experiments. A summary of these experiments is provided in Table~\ref{tab:Experiments}, and the corresponding client data distributions are detailed in Table~\ref{tab:client_distribution}.

We conduct eighteen experiments each on synthetic and real-world datasets, with the results presented in Section~\ref{sec:results}. For the real-world data, we follow the same preprocessing approach used for synthetic data (Section~\ref{sec:synthetic_data}). Specifically, we apply five different detrending techniques across three data distribution settings—generalized extreme value (GEV), log norm, and a mixed distribution—resulting in a total of eighteen experiments.

\begin{table}[t]
\caption{Experiments and detrending techniques used}
\resizebox{\columnwidth}{!}{%
\centering
\begin{tabular}{cccc}
\hline
\multicolumn{1}{l}{\multirow{2}{*}{\textbf{Exp Num}}} & \multirow{2}{*}{\textbf{Detrending technique}} & \multicolumn{2}{c}{\textbf{Client Data distribution}} \\ \cline{3-4} 
\multicolumn{1}{l}{}                                 &                                                & \multicolumn{1}{l}{\textbf{Gen-extreme}}      & \textbf{Log normal}     \\ \hline
\textbf{Exp 1}  & None                  & Yes & No  \\
\textbf{Exp 2}  & None                  & No  & Yes \\
\textbf{Exp 3}  & None                  & Yes & Yes \\
\textbf{Exp 4}  & Differencing          & Yes & No  \\
\textbf{Exp 5}  & Differencing          & No  & Yes \\
\textbf{Exp 6}  & Differencing          & Yes & Yes \\
\textbf{Exp 7}  & Moving average        & Yes & No  \\
\textbf{Exp 8}  & Moving average        & No  & Yes \\
\textbf{Exp 9}  & Moving average        & Yes & Yes \\
\textbf{Exp 10} & Subtracting mean      & Yes & No  \\
\textbf{Exp 11} & Subtracting mean      & No  & Yes \\
\textbf{Exp 12} & Subtracting mean      & Yes & Yes \\
\textbf{Exp 13} & Using linear model    & Yes & No  \\
\textbf{Exp 14} & Using linear model    & No  & Yes \\
\textbf{Exp 15} & Using linear model    & Yes & Yes \\
\textbf{Exp 16} & Using quadratic model & Yes & No  \\
\textbf{Exp 17} & Using quadratic model & No  & Yes \\
\textbf{Exp 18} & Using quadratic model & Yes & Yes \\ \hline
\end{tabular}
}
\label{tab:Experiments}
\end{table}

\begin{table}[!t]
\caption{Data distributions for various clients across different experiments}
\resizebox{\columnwidth}{!}{%
\begin{tabular}{lccc}
\hline
\multicolumn{4}{c}{\textbf{Data Distribution}}              \\ \cline{2-4} 
\multicolumn{1}{c}{\textbf{Client No}} &
  \textbf{\begin{tabular}[c]{@{}c@{}}Experiment \\ 1, 4, 7, 10, 13, 16\end{tabular}} &
  \textbf{\begin{tabular}[c]{@{}c@{}}Experiment\\ 2, 5, 8, 11, 14, 17\end{tabular}} &
  \textbf{\begin{tabular}[c]{@{}c@{}}Experiment\\ 3, 6, 9, 12, 15, 18\end{tabular}} \\ \hline
\textbf{Client 1}  & Gen-extreme & Log normal & Gen-extreme \\
\textbf{Client 2}  & Gen-extreme & Log normal & Gen-extreme \\
\textbf{Client 3}  & Gen-extreme & Log normal & Gen-extreme \\
\textbf{Client 4}  & Gen-extreme & Log normal & Gen-extreme \\
\textbf{Client 5}  & Gen-extreme & Log normal & Gen-extreme \\
\textbf{Client 6}  & Gen-extreme & Log normal & Log normal  \\
\textbf{Client 7}  & Gen-extreme & Log normal & Log normal  \\
\textbf{Client 8}  & Gen-extreme & Log normal & Log normal  \\
\textbf{Client 9}  & Gen-extreme & Log normal & Log normal  \\
\textbf{Client 10} & Gen-extreme & Log normal & Log normal  \\ \hline
\end{tabular}%
}
\label{tab:client_distribution}
\vspace{-10pt}
\end{table}

\section{Results and Discussion}
\label{sec:results}

\begin{table*}[!htb]
\caption{Centralized vs FL performance on synthetic data without detrending (Highlighted values show lower loss values and hence better performance)}
\setlength{\tabcolsep}{3pt}
\centering
\small
\begin{tabular}{@{}cccccccccc@{}}
\toprule
\multirow{2}{*}{\textbf{\begin{tabular}[c]{@{}c@{}}Training \\ approach\end{tabular}}} &
  \multicolumn{3}{c}{\textbf{Gen-extreme}} &
  \multicolumn{3}{c}{\textbf{Log norm}} &
  \multicolumn{3}{c}{\textbf{Both}} \\ \cmidrule(l){2-10} 
 &
  \textbf{MSE} &
  \textbf{RMSE} &
  \textbf{MAE} &
  \textbf{MSE} &
  \textbf{RMSE} &
  \textbf{MAE} &
  \textbf{MSE} &
  \textbf{RMSE} &
  \textbf{MAE} \\ \midrule
Centralized &
  \textbf{\begin{tabular}[c]{@{}c@{}}0.00273\\ (Exp 1)\end{tabular}} &
  \textbf{\begin{tabular}[c]{@{}c@{}}0.05145\\ (Exp 1)\end{tabular}} &
  \textbf{\begin{tabular}[c]{@{}c@{}}0.04740\\ (Exp 1)\end{tabular}} &
  \textbf{\begin{tabular}[c]{@{}c@{}}0.00589\\ (Exp 2)\end{tabular}} &
  \textbf{\begin{tabular}[c]{@{}c@{}}0.0755\\ (Exp 2)\end{tabular}} &
  \textbf{\begin{tabular}[c]{@{}c@{}}0.06945\\ (Exp 2)\end{tabular}} &
  \textbf{\begin{tabular}[c]{@{}c@{}}0.00904\\ (Exp 3)\end{tabular}} &
  \textbf{\begin{tabular}[c]{@{}c@{}}0.09098\\ (Exp 3)\end{tabular}} &
  \textbf{\begin{tabular}[c]{@{}c@{}}0.07258\\ (Exp 3)\end{tabular}} \\
FL &
  \begin{tabular}[c]{@{}c@{}}0.00844\\ (Exp 1)\end{tabular} &
  \begin{tabular}[c]{@{}c@{}}0.06809\\ (Exp 1)\end{tabular} &
  \begin{tabular}[c]{@{}c@{}}0.06006\\ (Exp 1)\end{tabular} &
  \begin{tabular}[c]{@{}c@{}}0.01583\\ (Exp 2)\end{tabular} &
  \begin{tabular}[c]{@{}c@{}}0.10854\\ (Exp 2)\end{tabular} &
  \begin{tabular}[c]{@{}c@{}}0.09814\\ (Exp 2)\end{tabular} &
  \begin{tabular}[c]{@{}c@{}}0.03115\\ (Exp 3)\end{tabular} &
  \begin{tabular}[c]{@{}c@{}}0.13781\\ (Exp 3)\end{tabular} &
  \begin{tabular}[c]{@{}c@{}}0.12718\\ (Exp 3)\end{tabular} \\ \bottomrule
\end{tabular}

\label{tab:central_vs_fed}
\end{table*}

\subsection{Results on Synthetic Dataset}

The results have been presented in two subsections: 1) the effect of non-linear data distributions on centralized and FL forecasting and 2) the effect of the detrending techniques on centralized and FL forecasting.

\subsubsection{Effect of Non-linear Time-series Data Distributions in Centralized and FL Setup}
\label{sec:effect_of_non_linear}

\begin{table*}[!h]
\caption{Centralized vs FL performance on synthetic data with different detrending techniques (Highlighted values show lower loss values and hence better performance.)}

\resizebox{\textwidth}{!}{%
\centering
\begin{tabular}{@{}ccccccccccc@{}}
\toprule
\multirow{2}{*}{\textbf{\begin{tabular}[c]{@{}c@{}}Training \\ approach\end{tabular}}} &
  \multirow{2}{*}{\textbf{Detrending technique}} &
  \multicolumn{3}{c}{\textbf{Gen-extreme}} &
  \multicolumn{3}{c}{\textbf{Log norm}} &
  \multicolumn{3}{c}{\textbf{Both}} \\ \cmidrule(l){3-11} 
 &
   &
  \textbf{MSE} &
  \textbf{RMSE} &
  \textbf{MAE} &
  \textbf{MSE} &
  \textbf{RMSE} &
  \textbf{MAE} &
  \textbf{MSE} &
  \textbf{RMSE} &
  \textbf{MAE} \\ \midrule
\multirow{5}{*}{Centralized} &
  Differencing &
  \begin{tabular}[c]{@{}c@{}}0.00062\\ (Exp 4)\end{tabular} &
  \begin{tabular}[c]{@{}c@{}}0.02364\\ (Exp 4)\end{tabular} &
  \begin{tabular}[c]{@{}c@{}}0.01819\\ (Exp 4)\end{tabular} &
  \begin{tabular}[c]{@{}c@{}}0.00092\\ (Exp 5)\end{tabular} &
  \begin{tabular}[c]{@{}c@{}}0.02850\\ (Exp 5)\end{tabular} &
  \begin{tabular}[c]{@{}c@{}}0.02329\\ (Exp 5)\end{tabular} &
  \begin{tabular}[c]{@{}c@{}}0.00640\\ (Exp 6)\end{tabular} &
  \begin{tabular}[c]{@{}c@{}}0.07655\\ (Exp 6)\end{tabular} &
  \begin{tabular}[c]{@{}c@{}}0.06235\\ (Exp 6)\end{tabular} \\
 &
  Moving average &
  \textbf{\begin{tabular}[c]{@{}c@{}}0.00060\\ (Exp 7)\end{tabular}} &
  \textbf{\begin{tabular}[c]{@{}c@{}}0.02234\\ (Exp 7)\end{tabular}} &
  \textbf{\begin{tabular}[c]{@{}c@{}}0.01793\\ (Exp 7)\end{tabular}} &
  \textbf{\begin{tabular}[c]{@{}c@{}}0.00071\\ (Exp 8)\end{tabular}} &
  \textbf{\begin{tabular}[c]{@{}c@{}}0.02486\\ (Exp 8)\end{tabular}} &
  \textbf{\begin{tabular}[c]{@{}c@{}}0.01956\\ (Exp 8)\end{tabular}} &
  \textbf{\begin{tabular}[c]{@{}c@{}}0.00414\\ (Exp 9)\end{tabular}} &
  \textbf{\begin{tabular}[c]{@{}c@{}}0.06074\\ (Exp 9)\end{tabular}} &
  \textbf{\begin{tabular}[c]{@{}c@{}}0.04824\\ (Exp 9)\end{tabular}} \\
 &
  Subtracting mean &
  \begin{tabular}[c]{@{}c@{}}0.00138\\ (Exp 10)\end{tabular} &
  \begin{tabular}[c]{@{}c@{}}0.03360\\ (Exp 10)\end{tabular} &
  \begin{tabular}[c]{@{}c@{}}0.02765\\ (Exp 10)\end{tabular} &
  \begin{tabular}[c]{@{}c@{}}0.00194\\ (Exp 11)\end{tabular} &
  \begin{tabular}[c]{@{}c@{}}0.04017\\ (Exp 11)\end{tabular} &
  \begin{tabular}[c]{@{}c@{}}0.03333\\ (Exp 11)\end{tabular} &
  \begin{tabular}[c]{@{}c@{}}0.00986\\ (Exp 12)\end{tabular} &
  \begin{tabular}[c]{@{}c@{}}0.09564\\ (Exp 12)\end{tabular} &
  \begin{tabular}[c]{@{}c@{}}0.07870\\ (Exp 12)\end{tabular} \\
 &
  Using linear model &
  \begin{tabular}[c]{@{}c@{}}0.00132\\ (Exp 13)\end{tabular} &
  \begin{tabular}[c]{@{}c@{}}0.03213\\ (Exp 13)\end{tabular} &
  \begin{tabular}[c]{@{}c@{}}0.02558\\ (Exp 13)\end{tabular} &
  \begin{tabular}[c]{@{}c@{}}0.00176\\ (Exp 14)\end{tabular} &
  \begin{tabular}[c]{@{}c@{}}0.03846\\ (Exp 14)\end{tabular} &
  \begin{tabular}[c]{@{}c@{}}0.03162\\ (Exp 14)\end{tabular} &
  \begin{tabular}[c]{@{}c@{}}0.01044\\ (Exp 15)\end{tabular} &
  \begin{tabular}[c]{@{}c@{}}0.09888\\ (Exp 15)\end{tabular} &
  \begin{tabular}[c]{@{}c@{}}0.08235\\ (Exp 15)\end{tabular} \\
 &
  Using quadratic model &
  \begin{tabular}[c]{@{}c@{}}0.00125\\ (Exp 16)\end{tabular} &
  \begin{tabular}[c]{@{}c@{}}0.03130\\ (Exp 16)\end{tabular} &
  \begin{tabular}[c]{@{}c@{}}0.02419\\ (Exp 16)\end{tabular} &
  \begin{tabular}[c]{@{}c@{}}0.00173\\ (Exp 17)\end{tabular} &
  \begin{tabular}[c]{@{}c@{}}0.03791\\ (Exp 17)\end{tabular} &
  \begin{tabular}[c]{@{}c@{}}0.03102\\ (Exp 17)\end{tabular} &
  \begin{tabular}[c]{@{}c@{}}0.01120\\ (Exp 18)\end{tabular} &
  \begin{tabular}[c]{@{}c@{}}0.10266\\ (Exp 18)\end{tabular} &
  \begin{tabular}[c]{@{}c@{}}0.08616\\ (Exp 18)\end{tabular} \\ \midrule
\multirow{5}{*}{FL} &
  Differencing &
  \textbf{\begin{tabular}[c]{@{}c@{}}0.00554\\ (Exp 4)\end{tabular}} &
  \textbf{\begin{tabular}[c]{@{}c@{}}0.04985\\ (Exp 4)\end{tabular}} &
  \textbf{\begin{tabular}[c]{@{}c@{}}0.04575\\ (Exp 4)\end{tabular}} &
  \textbf{\begin{tabular}[c]{@{}c@{}}0.01339\\ (Exp 5)\end{tabular}} &
  \textbf{\begin{tabular}[c]{@{}c@{}}0.09687\\ (Exp 5)\end{tabular}} &
  \textbf{\begin{tabular}[c]{@{}c@{}}0.08822\\ (Exp 5)\end{tabular}} &
  \textbf{\begin{tabular}[c]{@{}c@{}}0.01075\\ (Exp 6)\end{tabular}} &
  \textbf{\begin{tabular}[c]{@{}c@{}}0.07844\\ (Exp 6)\end{tabular}} &
  \textbf{\begin{tabular}[c]{@{}c@{}}0.07114\\ (Exp 6)\end{tabular}} \\
 &
  Moving average &
  \begin{tabular}[c]{@{}c@{}}0.00677\\ (Exp 7)\end{tabular} &
  \begin{tabular}[c]{@{}c@{}}0.06515\\ (Exp 7)\end{tabular} &
  \begin{tabular}[c]{@{}c@{}}0.05567\\ (Exp 7)\end{tabular} &
  \begin{tabular}[c]{@{}c@{}}0.0201\\ (Exp 8)\end{tabular} &
  \begin{tabular}[c]{@{}c@{}}0.12044\\ (Exp 8)\end{tabular} &
  \begin{tabular}[c]{@{}c@{}}0.10825\\ (Exp 8)\end{tabular} &
  \begin{tabular}[c]{@{}c@{}}0.0392\\ (Exp 9)\end{tabular} &
  \begin{tabular}[c]{@{}c@{}}0.14732\\ (Exp 9)\end{tabular} &
  \begin{tabular}[c]{@{}c@{}}0.12322\\ (Exp 9)\end{tabular} \\
 &
  Subtracting mean &
  \begin{tabular}[c]{@{}c@{}}0.0084\\ (Exp 10)\end{tabular} &
  \begin{tabular}[c]{@{}c@{}}0.06792\\ (Exp 10)\end{tabular} &
  \begin{tabular}[c]{@{}c@{}}0.0599\\ (Exp 10)\end{tabular} &
  \begin{tabular}[c]{@{}c@{}}0.01587\\ (Exp 11)\end{tabular} &
  \begin{tabular}[c]{@{}c@{}}0.10872\\ (Exp 11)\end{tabular} &
  \begin{tabular}[c]{@{}c@{}}0.09832\\ (Exp 11)\end{tabular} &
  \begin{tabular}[c]{@{}c@{}}0.03132\\ (Exp 12)\end{tabular} &
  \begin{tabular}[c]{@{}c@{}}0.13818\\ (Exp 12)\end{tabular} &
  \begin{tabular}[c]{@{}c@{}}0.12755\\ (Exp 12)\end{tabular} \\
 &
  Using linear model &
  \begin{tabular}[c]{@{}c@{}}0.00805\\ (Exp 13)\end{tabular} &
  \begin{tabular}[c]{@{}c@{}}0.0675\\ (Exp 13)\end{tabular} &
  \begin{tabular}[c]{@{}c@{}}0.05968\\ (Exp 13)\end{tabular} &
  \begin{tabular}[c]{@{}c@{}}0.01543\\ (Exp 14)\end{tabular} &
  \begin{tabular}[c]{@{}c@{}}0.10712\\ (Exp 14)\end{tabular} &
  \begin{tabular}[c]{@{}c@{}}0.09669\\ (Exp 14)\end{tabular} &
  \begin{tabular}[c]{@{}c@{}}0.0345\\ (Exp 15)\end{tabular} &
  \begin{tabular}[c]{@{}c@{}}0.14625\\ (Exp 15)\end{tabular} &
  \begin{tabular}[c]{@{}c@{}}0.136\\ (Exp 15)\end{tabular} \\
 &
  Using quadratic model &
  \begin{tabular}[c]{@{}c@{}}0.00778\\ (Exp 16)\end{tabular} &
  \begin{tabular}[c]{@{}c@{}}0.06782\\ (Exp 16)\end{tabular} &
  \begin{tabular}[c]{@{}c@{}}0.06034\\ (Exp 16)\end{tabular} &
  \begin{tabular}[c]{@{}c@{}}0.01532\\ (Exp 17)\end{tabular} &
  \begin{tabular}[c]{@{}c@{}}0.10682\\ (Exp 17)\end{tabular} &
  \begin{tabular}[c]{@{}c@{}}0.0964\\ (Exp 17)\end{tabular} &
  \begin{tabular}[c]{@{}c@{}}0.0351\\ (Exp 18)\end{tabular} &
  \begin{tabular}[c]{@{}c@{}}0.14858\\ (Exp 18)\end{tabular} &
  \begin{tabular}[c]{@{}c@{}}0.13868\\ (Exp 18)\end{tabular} \\ \bottomrule
\end{tabular}
}
    \label{tab:detrending_results}
    \vspace{-10pt}
\end{table*}

\begin{figure*}[]
    \centering
    \includegraphics[width=0.22\textwidth]{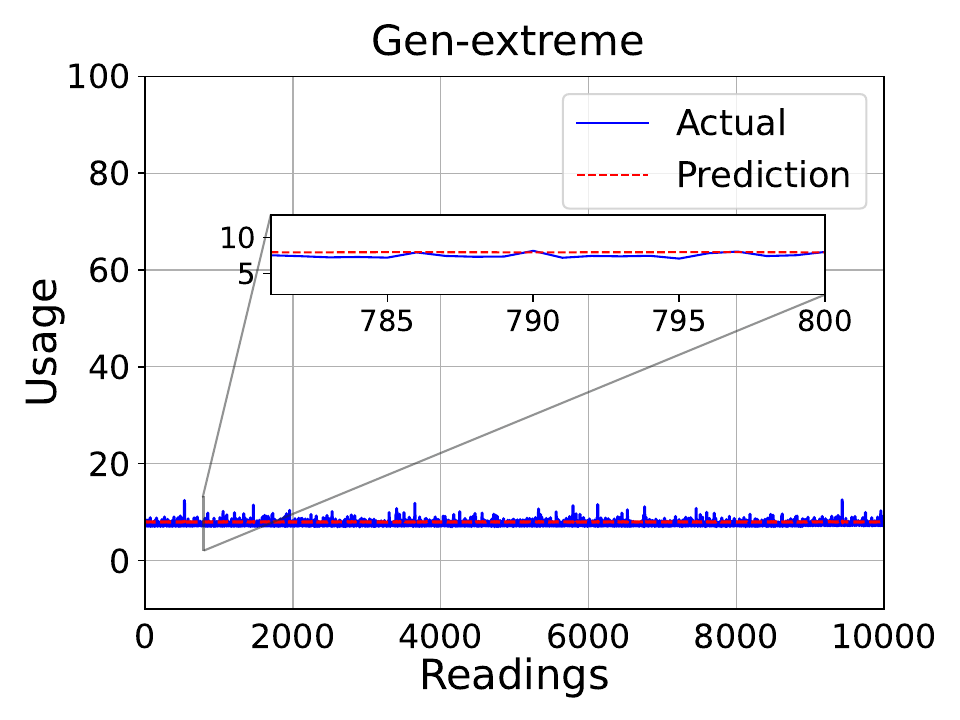}~
    \includegraphics[width=0.22\textwidth]{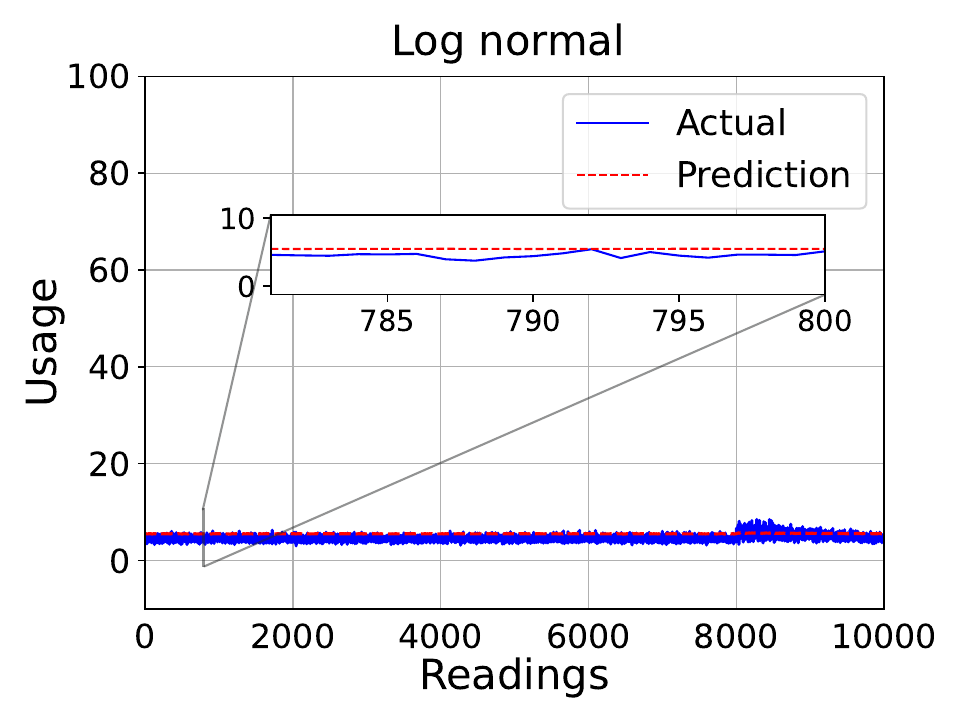}~
    \includegraphics[width=0.22\textwidth]{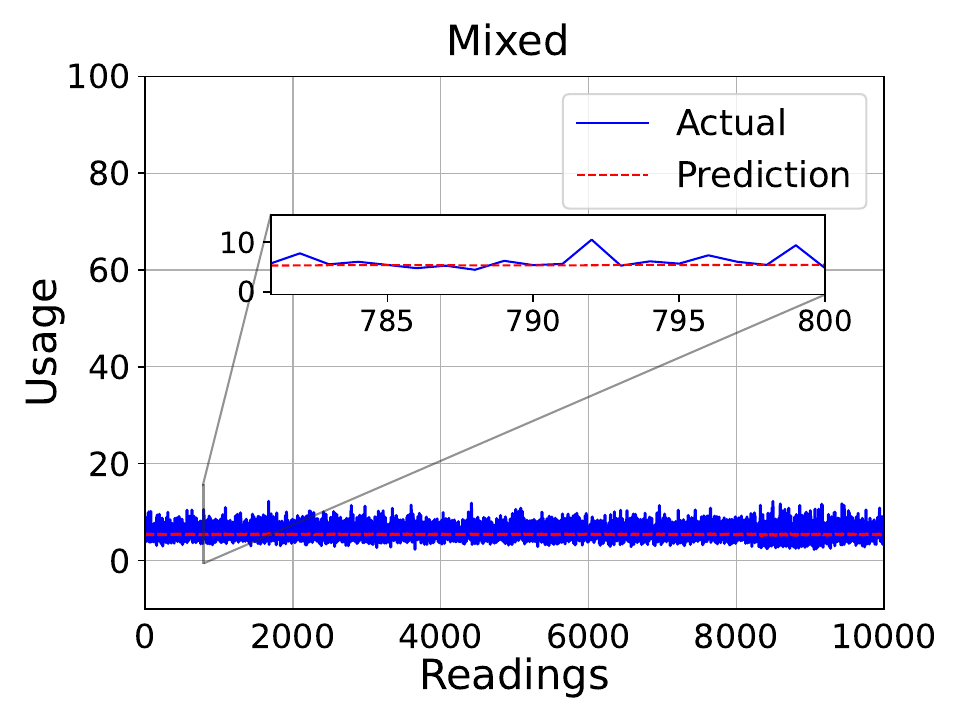}\\
    
    
    \includegraphics[width=0.22\textwidth]{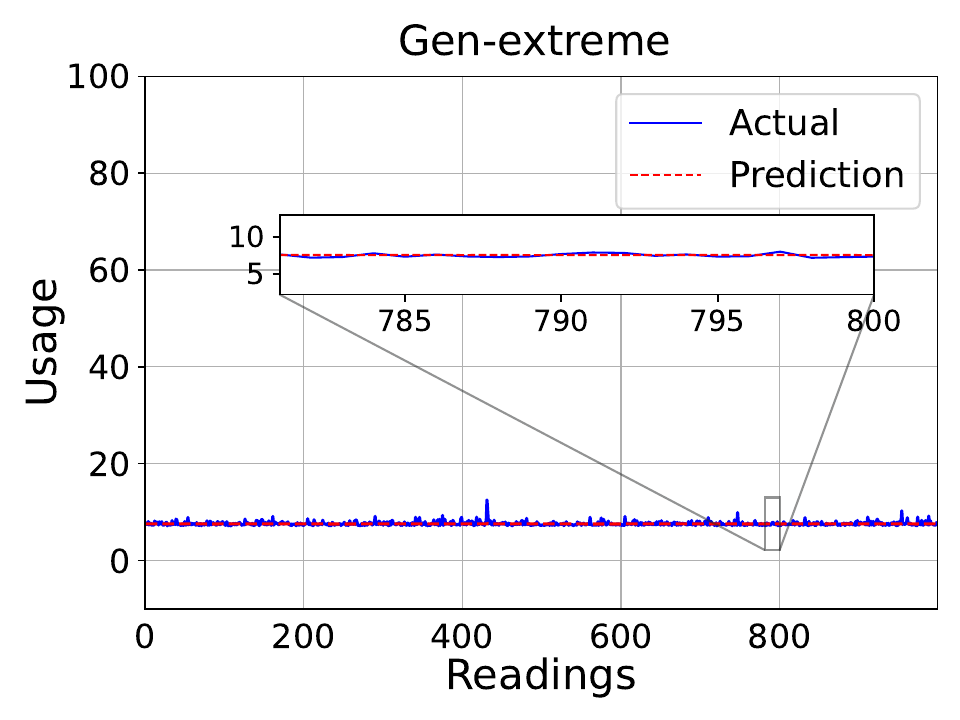}~
    \includegraphics[width=0.22\textwidth]{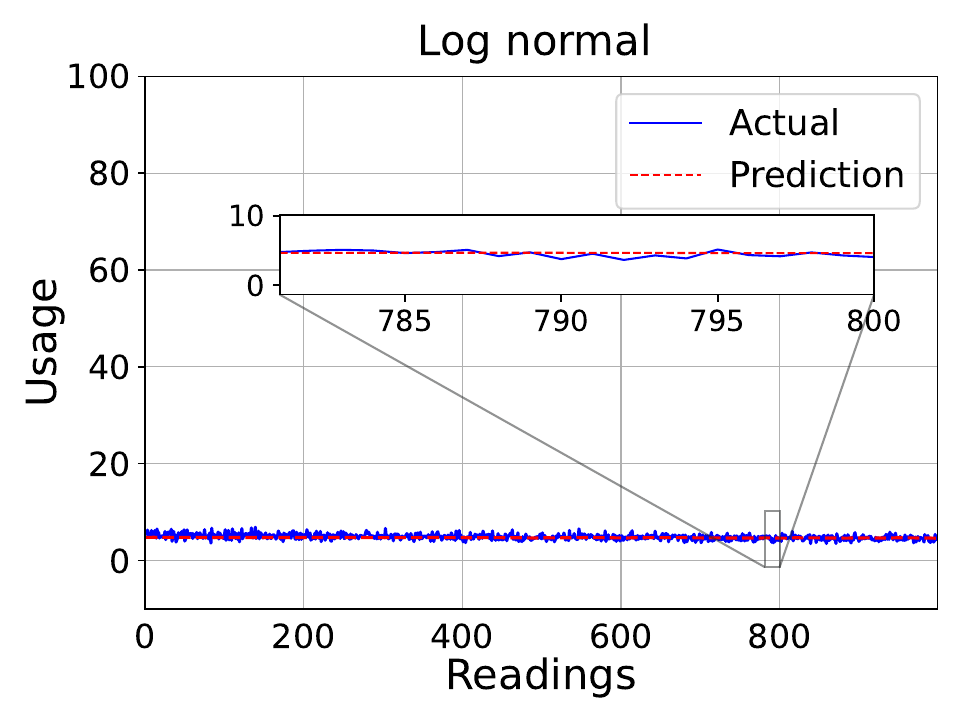}~
    \includegraphics[width=0.22\textwidth]{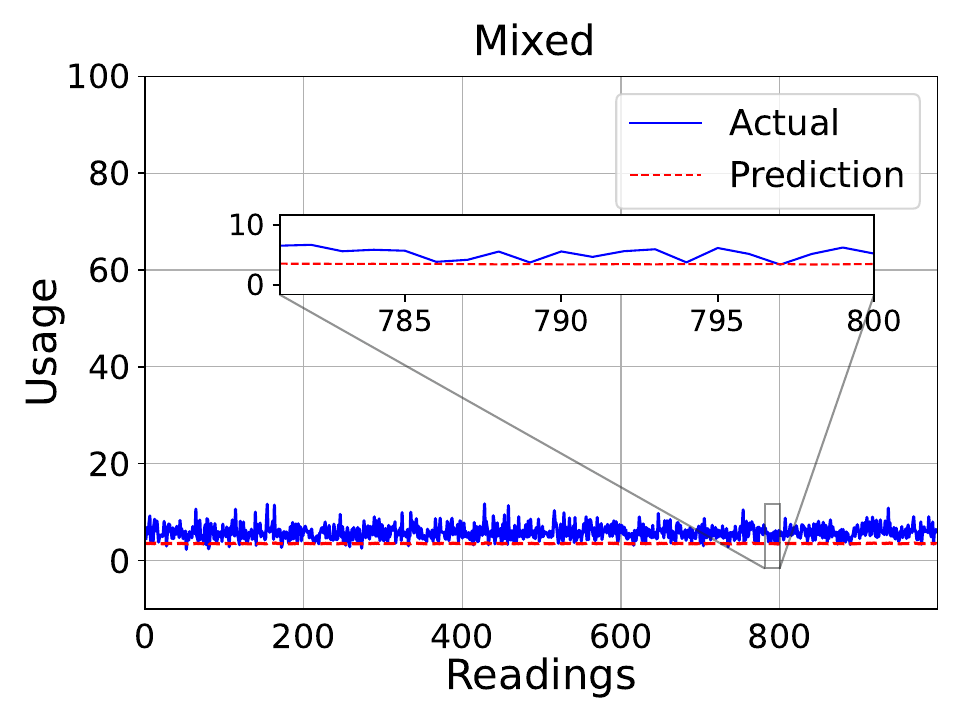}
    
    \caption{Model forecasting using both centralized (row 1) and FL (row 2) approach without any detrending for experiments 1, 2 and 3 conducted with synthetic data. 
    For the FL approach, results for only one of the clients are demonstrated. 
    }
    \label{fig:central_vs_fed_forecast}
    \vspace{-5pt}
\end{figure*}

\begin{figure*}[!htb]
        \includegraphics[width=0.196\textwidth]{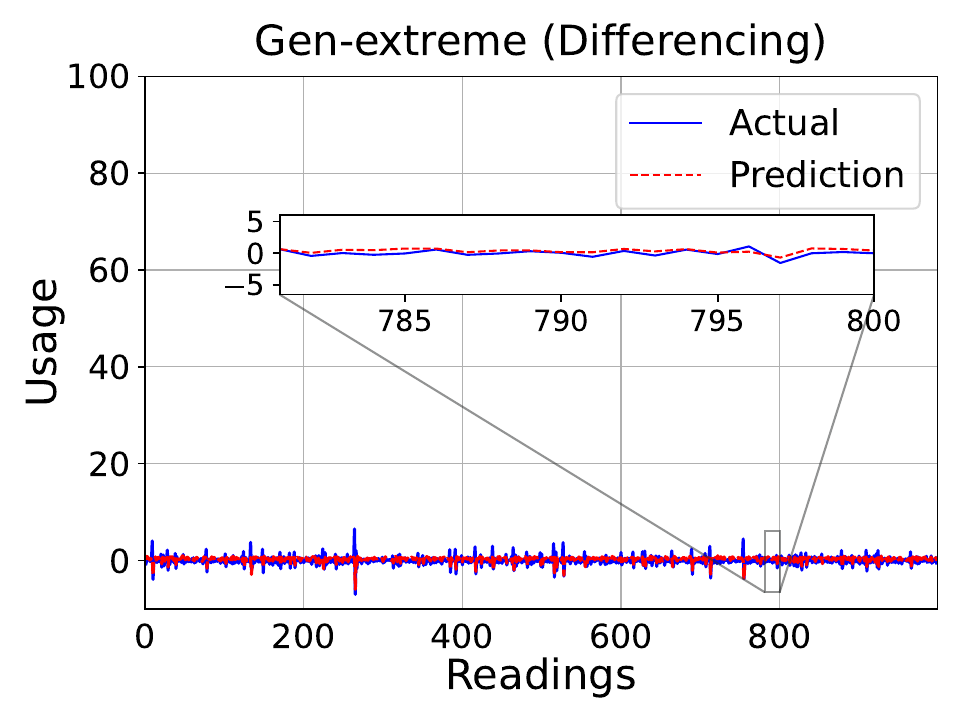}
        \includegraphics[width=0.196\textwidth]{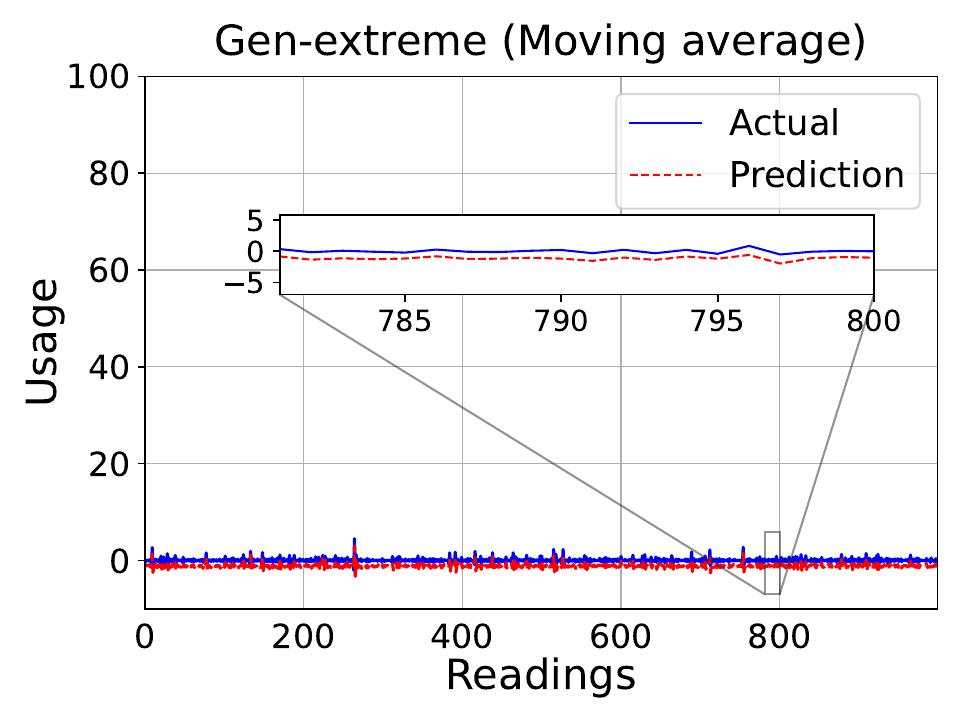}
        \includegraphics[width=0.196\textwidth]{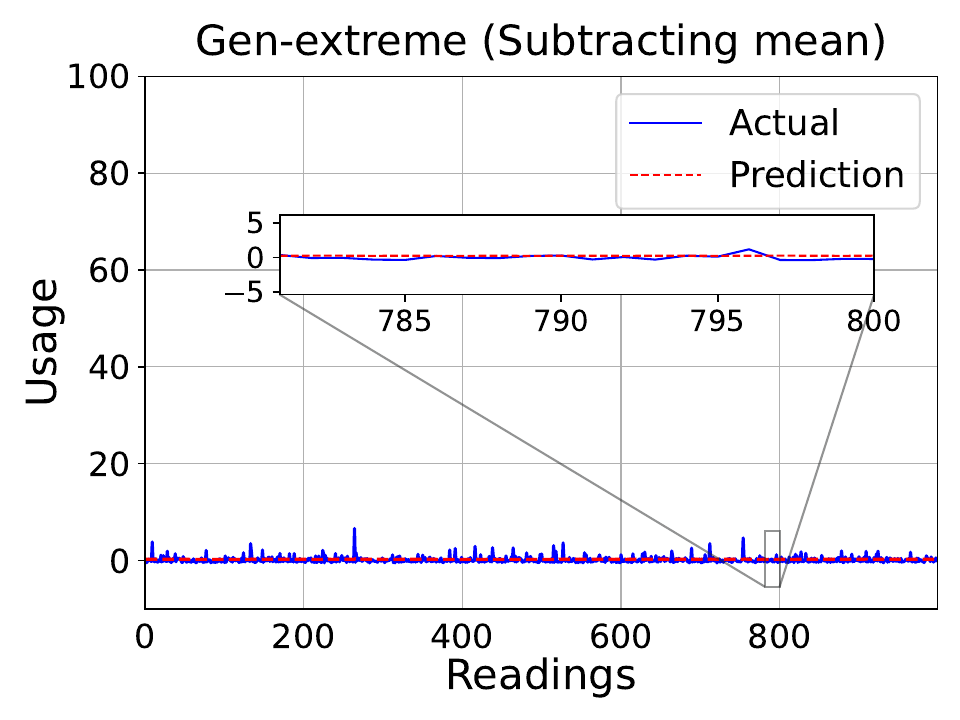}
        \includegraphics[width=0.196\textwidth]{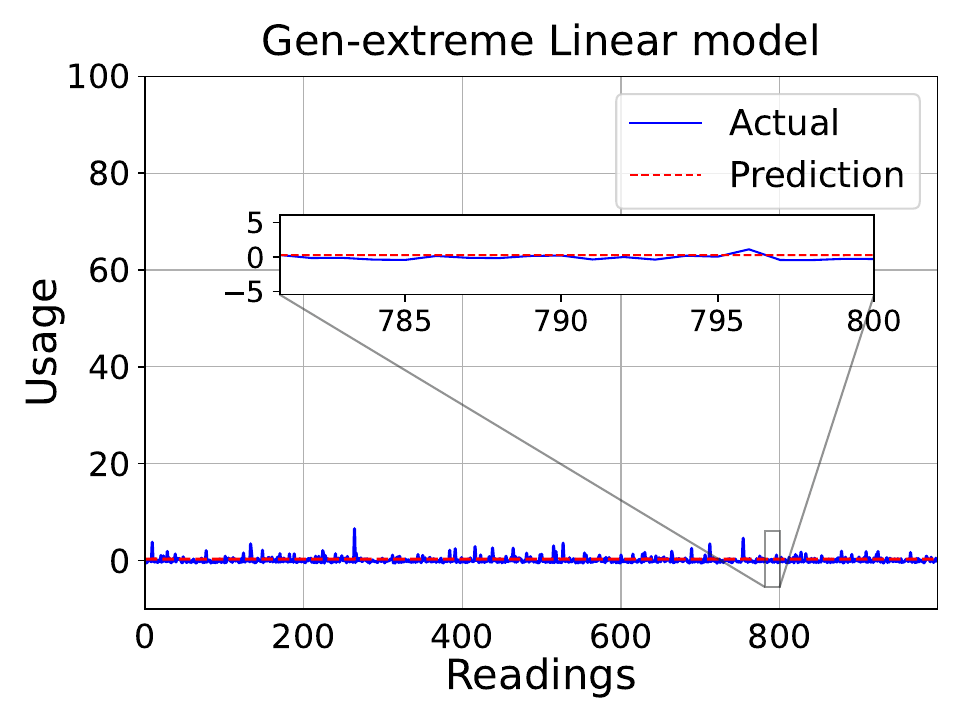}
        \includegraphics[width=0.196\textwidth]{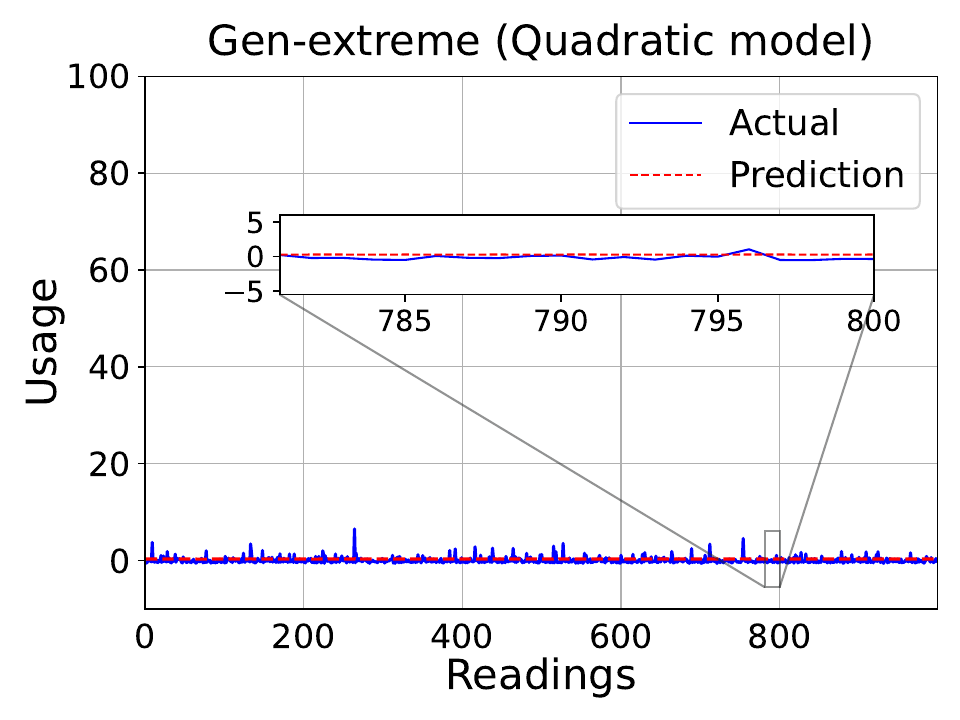}
    \caption{Model forecasting using FL approach for client 1, based on synthetic data following a gen-extreme distribution. The application of various detrending techniques is demonstrated, including differencing, moving average, mean removal, linear model, and quadratic model (from left to right).}
    \label{fig:c1_forecast}
    \vspace{-5pt}
\end{figure*}
\begin{figure*}[!htb]
        \includegraphics[width=0.196\textwidth]{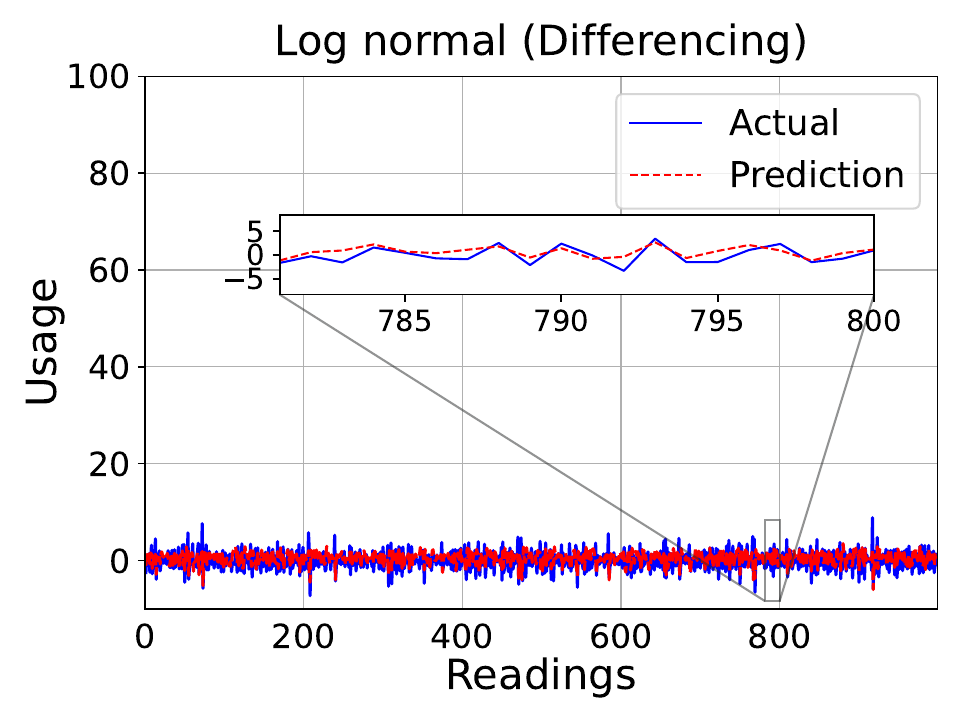}
        \includegraphics[width=0.196\textwidth]{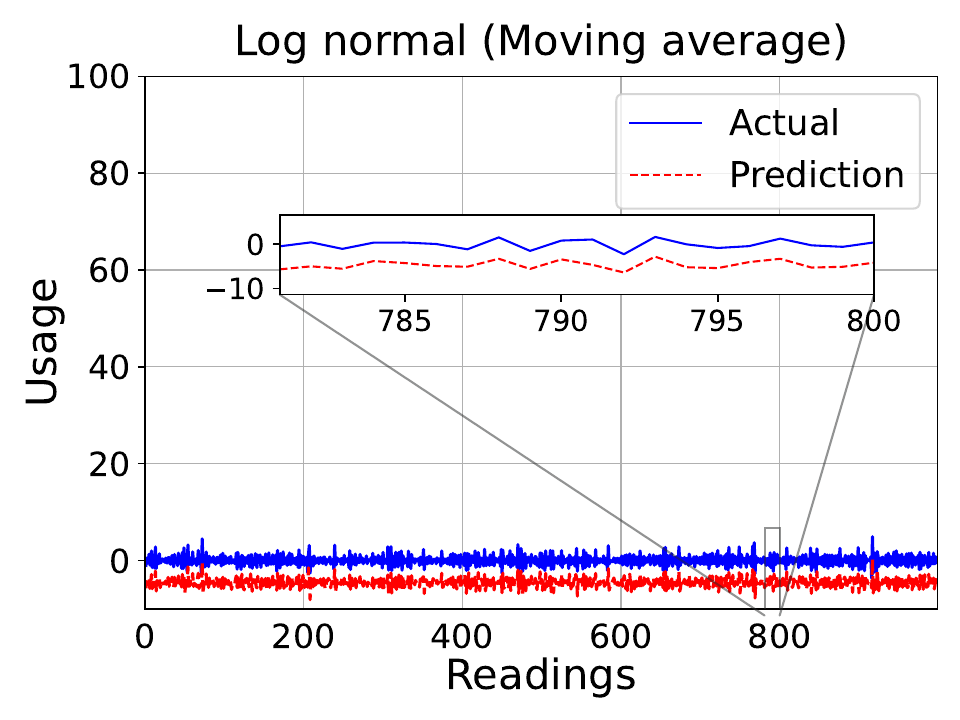}
        \includegraphics[width=0.196\textwidth]{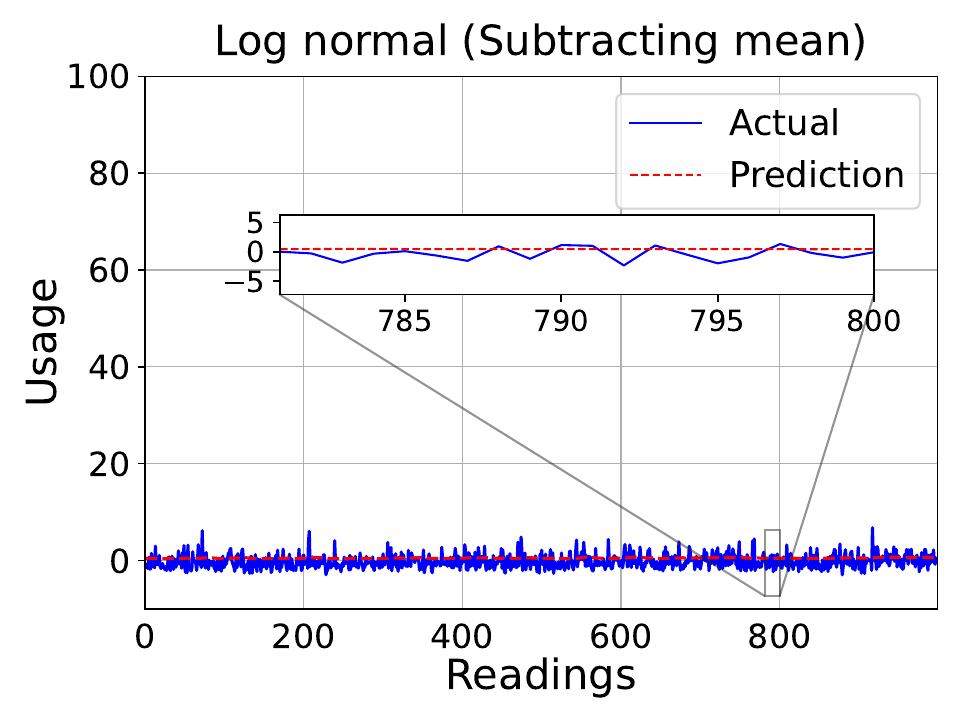}
        \includegraphics[width=0.196\textwidth]{images/new_synthetic/synthetic_exp_11_forecast_client_6_rnd_90.pdf}
        \includegraphics[width=0.196\textwidth]{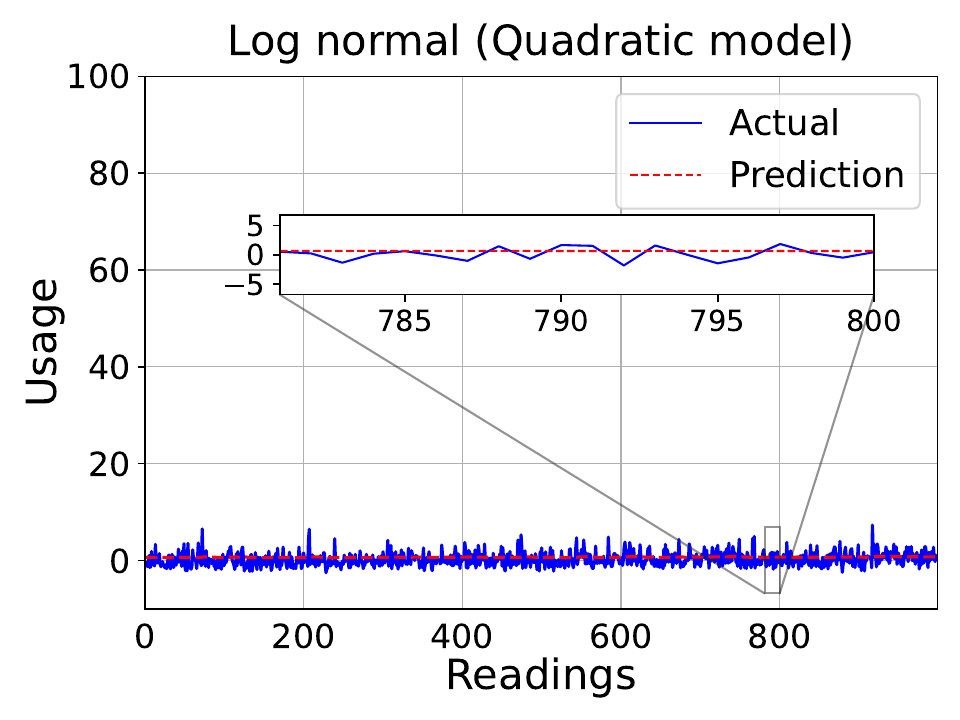}
    \caption{Model forecasting using FL approach for client 6, based on synthetic data following a log norm distribution. The application of various detrending techniques is demonstrated, including differencing, moving average, mean removal, linear model, and quadratic model (from left to right).}
    \label{fig:c6_forecast}
    \vspace{-10pt}
\end{figure*}

Three separate experiments were designed to assess the influence of non-linear data distributions on FL, focusing on generalized extreme value (gen-extreme), log normal (log norm), and a combination of both. In the first experiment, all clients used gen-extreme data; in the second, all clients used log norm data; and in the third, half of the clients had gen-extreme data while the other half used log norm.

To provide a baseline comparison, corresponding centralized training experiments were conducted for each FL setup. In these centralized settings, data from all clients were aggregated into a single dataset. For the first centralized experiment, this consisted of concatenated data from gen-extreme clients; the second used data from log norm clients; and the third used a mix from both distributions. No detrending was applied in any of these experiments.

The results from these non-linear distribution experiments (1, 2, and 3) highlight that data distribution significantly impacts model performance in both centralized and FL settings. In the centralized approach, experiment 3 (mixed distribution) had the highest average validation loss (measured using MSE, MAE, and RMSE), followed by experiment 2 (log norm), and then experiment 1 (gen-extreme). This ordering experiment 3 > experiment 2 > experiment 1 suggests that gen-extreme data, due to its lower variance and fewer extreme values, leads to less loss and better model performance. In contrast, the log norm distribution, characterized by higher variance, contributes more to the error.

Similarly, in the FL setup, the trend of validation loss remained the same: experiment 3 had the highest loss, followed by experiment 2, and the lowest was observed in experiment 1. The average MSE rose from 0.00844 in experiment 1 to 0.03115 in experiment 3, an increase of 269.07\% due to the inclusion of clients with log norm data. The clients’ average validation loss (MSE, MAE, RMSE) for different experiments are listed in Table \ref{tab:central_vs_fed}. In the FL setting, the elevated loss in experiment 3 is largely due to the influence of extreme values in the gen-extreme data. Conversely, models trained with log norm data, having fewer outliers, performed relatively better.

Notably, the centralized models consistently outperformed their FL counterparts across all data distributions. However, both centralised and FL models struggled to capture the unique characteristics of the data, failing to model the extremes in gen-extreme datasets and the variability in log norm datasets. These shortcomings are evident in the forecasts illustrated in Figure \ref{fig:central_vs_fed_forecast}.

\subsubsection{Effect of Detrending Techniques on Centralized and FL Setup}

To analyse the influence of detrending on model performance, we experimented with various techniques, including differencing, moving average, mean subtraction, and trend removal using linear and quadratic models. These methods were applied to three data distributions: gen-extreme, log norm, and a hybrid of both, resulting in a total of 15 experiments. Each experiment was conducted under both centralized and federated learning (FL) frameworks, making up 30 experiments in total.

In the centralized setup, as presented in Table \ref{tab:detrending_results}, the forecasting model yielded the least average validation loss when the moving average method was used, regardless of the data distribution. Moreover, in cases where clients exclusively used gen-extreme data (e.g., experiments 4 and 7), the variation in loss between differencing and moving average methods was negligible. This implies that, for gen-extreme data under centralized training, detrending may have a limited effect, likely because extreme values persist even after detrending.

In contrast, under the FL setup, the differencing technique proved to be the most effective at reducing forecasting error. It achieved the least average validation loss among all detrending methods, yielding an MSE of 0.00554 for clients with gen-extreme distributions and 0.01339 for those with log norm distributions. While the moving average technique also helped reduce loss to some extent, it was less effective than differencing. Other methods, such as subtracting the mean, or removing trends using linear or quadratic models, showed little to no improvement in model performance.

Forecast visualizations for gen-extreme and log norm distributions are shown in Figures \ref{fig:c1_forecast} and \ref{fig:c6_forecast}, illustrating the comparative impact of each detrending technique. Displayed left to right—differencing, moving average, subtracting mean, linear model, and quadratic model, the figures demonstrate that choosing an effective detrending method can reduce forecasting errors, with differencing consistently outperforming the others across all distributions.


\subsection{Results on Real-world Dataset}
\begin{table*}[]
\caption{Centralized vs FL performance on Ausgrid dataset without detrending (Highlighted values show lower loss values and hence better performance)}

\setlength{\tabcolsep}{3pt}
\centering
\small
\begin{tabular}{@{}cccccccccc@{}}
\toprule
\multirow{2}{*}{\textbf{\begin{tabular}[c]{@{}c@{}}Training \\ approach\end{tabular}}} &
  \multicolumn{3}{c}{\textbf{Gen-extreme}} &
  \multicolumn{3}{c}{\textbf{Log norm}} &
  \multicolumn{3}{c}{\textbf{Both}} \\ \cmidrule(l){2-10} 
 &
  \textbf{MSE} &
  \textbf{RMSE} &
  \textbf{MAE} &
  \textbf{MSE} &
  \textbf{RMSE} &
  \textbf{MAE} &
  \textbf{MSE} &
  \textbf{RMSE} &
  \textbf{MAE} \\ \midrule
Centralized &
  \textbf{\begin{tabular}[c]{@{}c@{}}0.0000069\\ (Exp 1)\end{tabular}} &
  \textbf{\begin{tabular}[c]{@{}c@{}}0.0078302\\ (Exp 1)\end{tabular}} &
  \textbf{\begin{tabular}[c]{@{}c@{}}0.0068493\\ (Exp 1)\end{tabular}} &
  \textbf{\begin{tabular}[c]{@{}c@{}}0.0003645\\ (Exp 2)\end{tabular}} &
  \textbf{\begin{tabular}[c]{@{}c@{}}0.0174808\\ (Exp 2)\end{tabular}} &
  \textbf{\begin{tabular}[c]{@{}c@{}}0.0153261\\ (Exp 2)\end{tabular}} &
  \textbf{\begin{tabular}[c]{@{}c@{}}0.0001674\\ (Exp 3)\end{tabular}} &
  \textbf{\begin{tabular}[c]{@{}c@{}}0.0128746\\ (Exp 3)\end{tabular}} &
  \textbf{\begin{tabular}[c]{@{}c@{}}0.0105933\\ (Exp 3)\end{tabular}} \\
FL &
  \begin{tabular}[c]{@{}c@{}}0.00286\\ (Exp 1)\end{tabular} &
  \begin{tabular}[c]{@{}c@{}}0.02741\\ (Exp 1)\end{tabular} &
  \begin{tabular}[c]{@{}c@{}}0.02519\\ (Exp 1)\end{tabular} &
  \begin{tabular}[c]{@{}c@{}}0.00477\\ (Exp 2)\end{tabular} &
  \begin{tabular}[c]{@{}c@{}}0.03886\\ (Exp 2)\end{tabular} &
  \begin{tabular}[c]{@{}c@{}}0.03511\\ (Exp 2)\end{tabular} &
  \begin{tabular}[c]{@{}c@{}}0.00374\\ (Exp 3)\end{tabular} &
  \begin{tabular}[c]{@{}c@{}}0.0346\\ (Exp 3)\end{tabular} &
  \begin{tabular}[c]{@{}c@{}}0.03154\\ (Exp 3)\end{tabular} \\ \bottomrule
\end{tabular}

\label{tab:ausgrid_central_vs_fed}
\end{table*}

\begin{table*}[]
\caption{Centralized vs FL performance on Ausgrid dataset with different detrending techniques (Highlighted values show lower loss values and hence better performance.)}
\vspace{-4pt}
\resizebox{\textwidth}{!}{%
\centering
\begin{tabular}{@{}ccccccccccc@{}}
\toprule
\multirow{2}{*}{\textbf{\begin{tabular}[c]{@{}c@{}}Training \\ approach\end{tabular}}} &
  \multirow{2}{*}{\textbf{Detrending technique}} &
  \multicolumn{3}{c}{\textbf{Gen-extreme}} &
  \multicolumn{3}{c}{\textbf{Log norm}} &
  \multicolumn{3}{c}{\textbf{Both}} \\ \cmidrule(l){3-11} 
 &
   &
  \textbf{MSE} &
  \textbf{RMSE} &
  \textbf{MAE} &
  \textbf{MSE} &
  \textbf{RMSE} &
  \textbf{MAE} &
  \textbf{MSE} &
  \textbf{RMSE} &
  \textbf{MAE} \\ \midrule
\multirow{5}{*}{Centralized} &
  Differencing &
  \begin{tabular}[c]{@{}c@{}}0.0000018\\ (Exp 4)\end{tabular} &
  \textbf{\begin{tabular}[c]{@{}c@{}}0.0052\\ (Exp 4)\end{tabular}} &
  \textbf{\begin{tabular}[c]{@{}c@{}}0.00431\\ (Exp 4)\end{tabular}} &
  \begin{tabular}[c]{@{}c@{}}0.000281\\ (Exp 5)\end{tabular} &
  \textbf{\begin{tabular}[c]{@{}c@{}}0.0125\\ (Exp 5)\end{tabular}} &
  \textbf{\begin{tabular}[c]{@{}c@{}}0.0094\\ (Exp 5)\end{tabular}} &
  \begin{tabular}[c]{@{}c@{}}0.000221\\ (Exp 6)\end{tabular} &
  \textbf{\begin{tabular}[c]{@{}c@{}}0.0105\\ (Exp 6)\end{tabular}} &
  \textbf{\begin{tabular}[c]{@{}c@{}}0.00782\\ (Exp 6)\end{tabular}} \\
 &
  Moving average &
  \textbf{\begin{tabular}[c]{@{}c@{}}0.0000011\\ (Exp 7)\end{tabular}} &
  \begin{tabular}[c]{@{}c@{}}0.0077\\ (Exp 7)\end{tabular} &
  \begin{tabular}[c]{@{}c@{}}0.00694\\ (Exp 7)\end{tabular} &
  \begin{tabular}[c]{@{}c@{}}0.000237\\ (Exp 8)\end{tabular} &
  \begin{tabular}[c]{@{}c@{}}0.0167\\ (Exp 8)\end{tabular} &
  \begin{tabular}[c]{@{}c@{}}0.0146\\ (Exp 8)\end{tabular} &
  \begin{tabular}[c]{@{}c@{}}0.000172\\ (Exp 9)\end{tabular} &
  \begin{tabular}[c]{@{}c@{}}0.0152\\ (Exp 9)\end{tabular} &
  \begin{tabular}[c]{@{}c@{}}0.01267\\ (Exp 9)\end{tabular} \\
 &
  Subtracting mean &
  \begin{tabular}[c]{@{}c@{}}0.0000114\\ (Exp 10)\end{tabular} &
  \begin{tabular}[c]{@{}c@{}}0.0079\\ (Exp 10)\end{tabular} &
  \begin{tabular}[c]{@{}c@{}}0.00688\\ (Exp 10)\end{tabular} &
  \begin{tabular}[c]{@{}c@{}}0.000279\\ (Exp 11)\end{tabular} &
  \begin{tabular}[c]{@{}c@{}}0.0150\\ (Exp 11)\end{tabular} &
  \begin{tabular}[c]{@{}c@{}}0.0128\\ (Exp 11)\end{tabular} &
  \begin{tabular}[c]{@{}c@{}}0.000171\\ (Exp 12)\end{tabular} &
  \begin{tabular}[c]{@{}c@{}}0.0122\\ (Exp 12)\end{tabular} &
  \begin{tabular}[c]{@{}c@{}}0.01003\\ (Exp 12)\end{tabular} \\
 &
  Using linear model &
  \begin{tabular}[c]{@{}c@{}}0.0000068\\ (Exp 13)\end{tabular} &
  \begin{tabular}[c]{@{}c@{}}0.0072\\ (Exp 13)\end{tabular} &
  \begin{tabular}[c]{@{}c@{}}0.00626\\ (Exp 13)\end{tabular} &
  \textbf{\begin{tabular}[c]{@{}c@{}}0.000202\\ (Exp 14)\end{tabular}} &
  \begin{tabular}[c]{@{}c@{}}0.0131\\ (Exp 14)\end{tabular} &
  \begin{tabular}[c]{@{}c@{}}0.0109\\ (Exp 14)\end{tabular} &
  \textbf{\begin{tabular}[c]{@{}c@{}}0.000156\\ (Exp 15)\end{tabular}} &
  \begin{tabular}[c]{@{}c@{}}0.0115\\ (Exp 15)\end{tabular} &
  \begin{tabular}[c]{@{}c@{}}0.00948\\ (Exp 15)\end{tabular} \\
 &
  Using quadratic model &
  \begin{tabular}[c]{@{}c@{}}0.0000098\\ (Exp 16)\end{tabular} &
  \begin{tabular}[c]{@{}c@{}}0.0074\\ (Exp 16)\end{tabular} &
  \begin{tabular}[c]{@{}c@{}}0.00634\\ (Exp 16)\end{tabular} &
  \begin{tabular}[c]{@{}c@{}}0.000231\\ (Exp 17)\end{tabular} &
  \begin{tabular}[c]{@{}c@{}}0.0141\\ (Exp 17)\end{tabular} &
  \begin{tabular}[c]{@{}c@{}}0.0119\\ (Exp 17)\end{tabular} &
  \begin{tabular}[c]{@{}c@{}}0.000159\\ (Exp 18)\end{tabular} &
  \begin{tabular}[c]{@{}c@{}}0.0119\\ (Exp 18)\end{tabular} &
  \begin{tabular}[c]{@{}c@{}}0.00994\\ (Exp 18)\end{tabular} \\ \midrule
\multirow{5}{*}{FL} &
  Differencing &
  \begin{tabular}[c]{@{}c@{}}0.00629\\ (Exp 4)\end{tabular} &
  \begin{tabular}[c]{@{}c@{}}0.0496 \\ (Exp 4)\end{tabular} &
  \begin{tabular}[c]{@{}c@{}}0.0465 \\ (Exp 4)\end{tabular} &
  \begin{tabular}[c]{@{}c@{}}0.0098 \\ (Exp 5)\end{tabular} &
  \begin{tabular}[c]{@{}c@{}}0.06336 \\ (Exp 5)\end{tabular} &
  \begin{tabular}[c]{@{}c@{}}0.05804 \\ (Exp 5)\end{tabular} &
  \begin{tabular}[c]{@{}c@{}}0.00779 \\ (Exp 6)\end{tabular} &
  \begin{tabular}[c]{@{}c@{}}0.05500 \\ (Exp 6)\end{tabular} &
  \begin{tabular}[c]{@{}c@{}}0.0507 \\ (Exp 6)\end{tabular} \\
 &
  Moving average &
  \begin{tabular}[c]{@{}c@{}}0.06001 \\ (Exp 7)\end{tabular} &
  \begin{tabular}[c]{@{}c@{}}0.1387\\ (Exp 7)\end{tabular} &
  \begin{tabular}[c]{@{}c@{}}0.1239 \\ (Exp 7)\end{tabular} &
  \begin{tabular}[c]{@{}c@{}}0.0087 \\ (Exp 8)\end{tabular} &
  \begin{tabular}[c]{@{}c@{}}0.07227 \\ (Exp 8)\end{tabular} &
  \begin{tabular}[c]{@{}c@{}}0.06941 \\ (Exp 8)\end{tabular} &
  \begin{tabular}[c]{@{}c@{}}0.04147 \\ (Exp 9)\end{tabular} &
  \begin{tabular}[c]{@{}c@{}}0.10584 \\ (Exp 9)\end{tabular} &
  \begin{tabular}[c]{@{}c@{}}0.0938 \\ (Exp 9)\end{tabular} \\
 &
  Subtracting mean &
  \begin{tabular}[c]{@{}c@{}}0.00282 \\ (Exp 10)\end{tabular} &
  \begin{tabular}[c]{@{}c@{}}0.0274 \\ (Exp 10)\end{tabular} &
  \begin{tabular}[c]{@{}c@{}}0.0251 \\ (Exp 10)\end{tabular} &
  \textbf{\begin{tabular}[c]{@{}c@{}}0.0047 \\ (Exp 11)\end{tabular}} &
  \textbf{\begin{tabular}[c]{@{}c@{}}0.03841 \\ (Exp 11)\end{tabular}} &
  \textbf{\begin{tabular}[c]{@{}c@{}}0.03463 \\ (Exp 11)\end{tabular}} &
  \begin{tabular}[c]{@{}c@{}}0.00373 \\ (Exp 12)\end{tabular} &
  \begin{tabular}[c]{@{}c@{}}0.03455 \\ (Exp 12)\end{tabular} &
  \begin{tabular}[c]{@{}c@{}}0.0315 \\ (Exp 12)\end{tabular} \\
 &
  Using linear model &
  \begin{tabular}[c]{@{}c@{}}0.00287 \\ (Exp 13)\end{tabular} &
  \begin{tabular}[c]{@{}c@{}}0.0265 \\ (Exp 13)\end{tabular} &
  \begin{tabular}[c]{@{}c@{}}0.0244\\ (Exp 13)\end{tabular} &
  \begin{tabular}[c]{@{}c@{}}0.0051 \\ (Exp 14)\end{tabular} &
  \begin{tabular}[c]{@{}c@{}}0.03881 \\ (Exp 14)\end{tabular} &
  \begin{tabular}[c]{@{}c@{}}0.03511 \\ (Exp 14)\end{tabular} &
  \begin{tabular}[c]{@{}c@{}}0.00382 \\ (Exp 15)\end{tabular} &
  \begin{tabular}[c]{@{}c@{}}0.03393 \\ (Exp 15)\end{tabular} &
  \begin{tabular}[c]{@{}c@{}}0.0309 \\ (Exp 15)\end{tabular} \\
 &
  Using quadratic model &
  \textbf{\begin{tabular}[c]{@{}c@{}}0.00276 \\ (Exp 16)\end{tabular}} &
  \textbf{\begin{tabular}[c]{@{}c@{}}0.0262 \\ (Exp 16)\end{tabular}} &
  \textbf{\begin{tabular}[c]{@{}c@{}}0.0241 \\ (Exp 16)\end{tabular}} &
  \begin{tabular}[c]{@{}c@{}}0.0049 \\ (Exp 17)\end{tabular} &
  \begin{tabular}[c]{@{}c@{}}0.03847 \\ (Exp 17)\end{tabular} &
  \begin{tabular}[c]{@{}c@{}}0.03481 \\ (Exp 17)\end{tabular} &
  \textbf{\begin{tabular}[c]{@{}c@{}}0.00366 \\ (Exp 18)\end{tabular}} &
  \textbf{\begin{tabular}[c]{@{}c@{}}0.03359 \\ (Exp 18)\end{tabular}} &
  \textbf{\begin{tabular}[c]{@{}c@{}}0.0307 \\ (Exp 18)\end{tabular}} \\ \bottomrule
\end{tabular}
}

    \label{tab:ausgrid_detrending_results}
    \vspace{-10pt}
\end{table*}

This subsection discusses the experimental results for the Ausgrid dataset. 

\subsubsection{Effect of Non-linear Time-series Data Distributions in
Centralized and FL Setup}
\begin{figure*}[!htbp]
    \centering
    \includegraphics[width=0.22\textwidth]{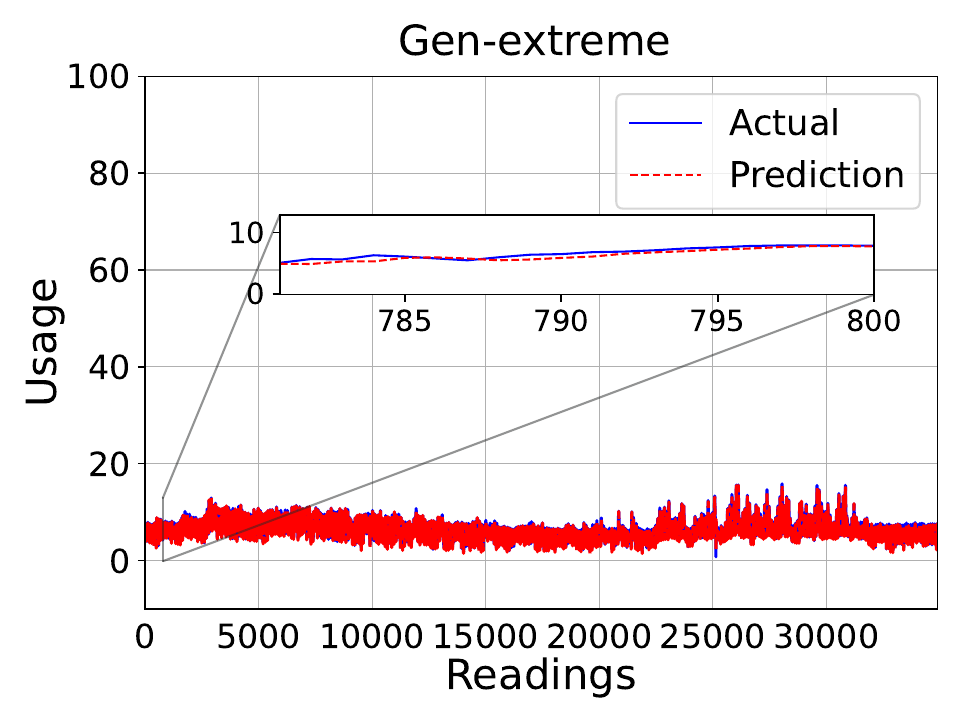}~
    \includegraphics[width=0.22\textwidth]{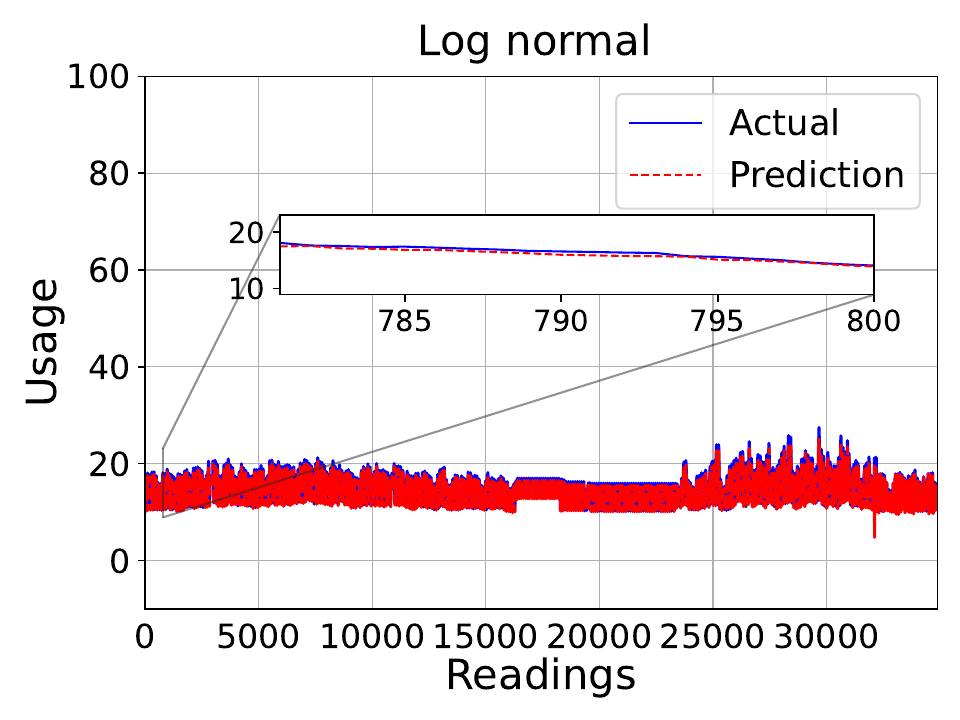}~
    \includegraphics[width=0.22\textwidth]{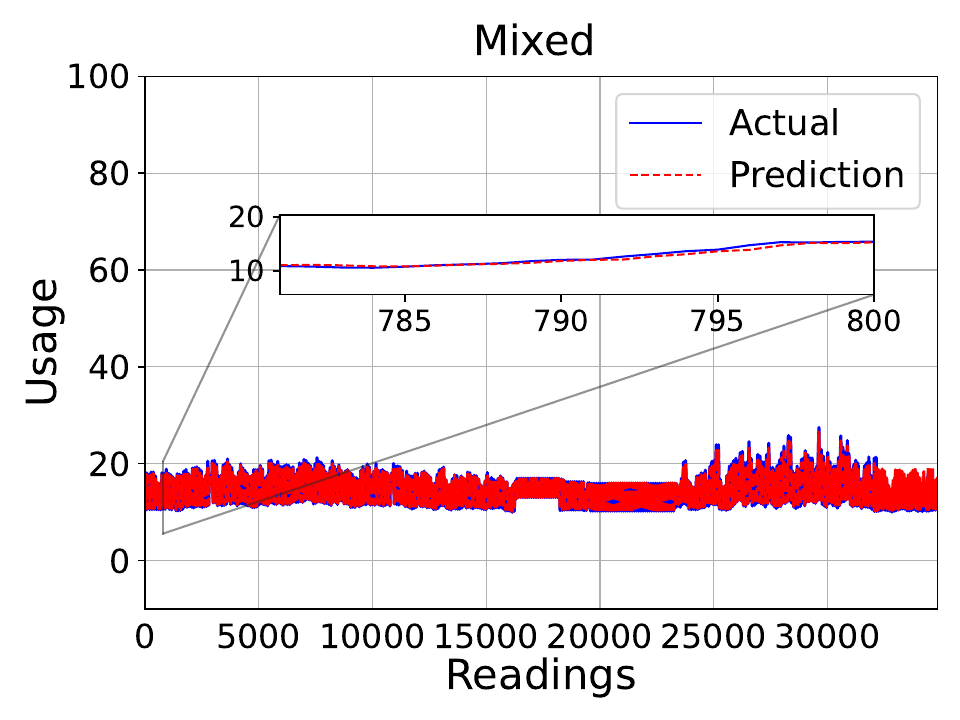}\\
    
    
    \includegraphics[width=0.22\textwidth]{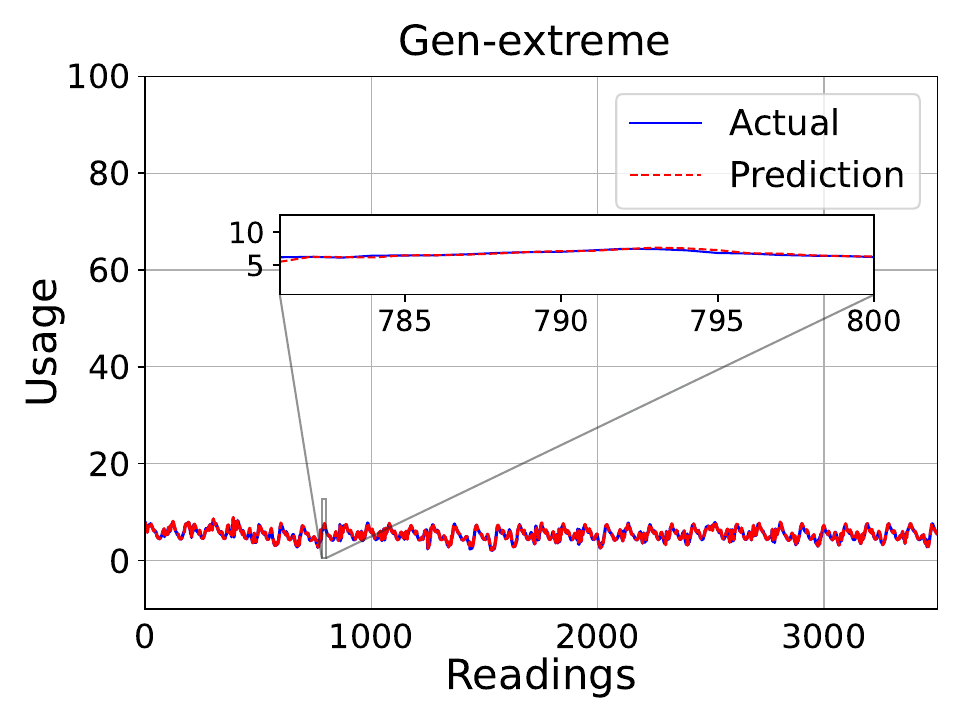}~
    \includegraphics[width=0.22\textwidth]{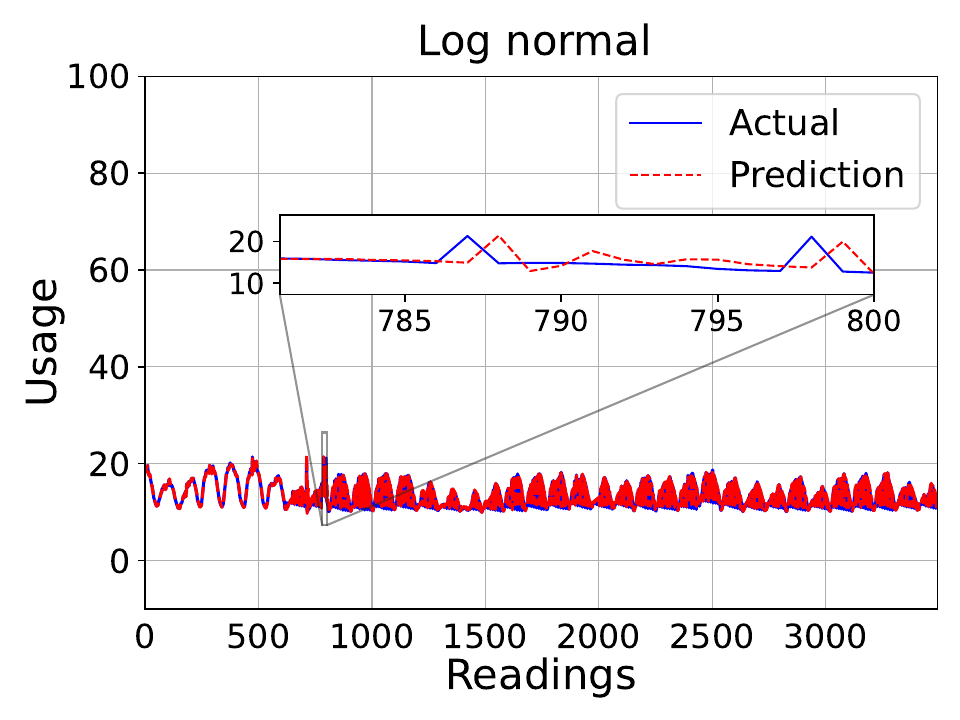}~
    \includegraphics[width=0.22\textwidth]{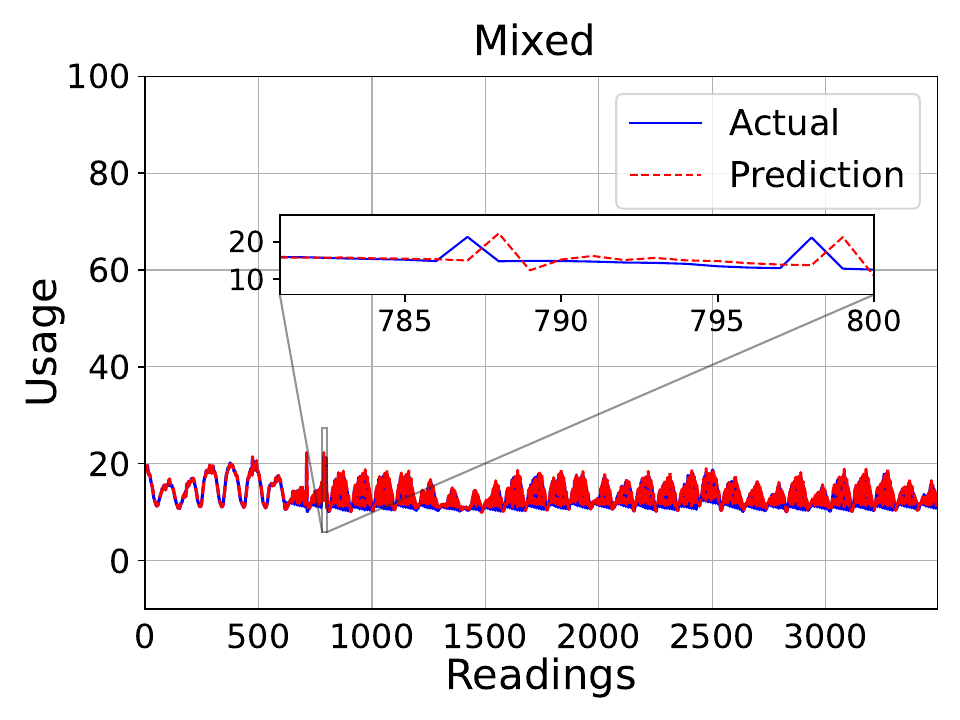}
    
    \caption{Model forecasting using both centralized (row 1) and FL (row 2) approach without any detrending for experiments 1, 2 and 3 conducted with Ausgrid data. 
    For the FL approach, results for only one of the clients are demonstrated.}
    \label{fig:ausgrid_central_vs_fed_forecast}
    \vspace{-5pt}
\end{figure*}

We evaluated the impact of three non-linear data distributions, generalized extreme value (gen-extreme), log norm, and a mixed distribution (combining gen-extreme and log norm), in both centralized and FL settings, without applying any detrending. As shown in Table~\ref{tab:ausgrid_central_vs_fed}, the centralized models consistently achieved lower average MSE, RMSE, and MAE compared to their FL counterparts. Figure~\ref{fig:ausgrid_central_vs_fed_forecast} further illustrates that forecasting errors in the centralized setup were marginally lower than those observed in FL.

Among the three experiments, the average validation loss (MSE, MAE, RMSE) was highest for the log norm distribution (Experiment 2), followed by the mixed distribution (Experiment 3), and lowest for the gen-extreme distribution (Experiment 1). This trend was consistent across both centralized and FL setups, suggesting that clients with log norm distributed data contribute more significantly to overall prediction errors. Notably, the average MSE increased from 0.00286 in Experiment 1 to 0.00374 in Experiment 3, with a 30.77\% rise due to the inclusion of log norm data clients.

\subsubsection{Effect of Detrending Techniques on Centralized and FL Setup}
Following the methodology used with synthetic data, we applied five detrending techniques: differencing, moving average, mean subtraction, and linear and quadratic trend removal to the Ausgrid dataset. This resulted in a total of 30 experiments: 15 under centralized training and 15 under the FL setup, as summarized in Table~\ref{tab:ausgrid_detrending_results}.

In the centralized setup, the differencing technique consistently led to reductions in average RMSE and MAE across all three distribution types (gen-extreme, log norm, and mixed). For log norm and mixed distributions, linear detrending further reduced average MSE, while the moving average technique proved most effective for gen-extreme distributions.

Under the FL setup, quadratic detrending yielded the best results for gen-extreme (MSE = 0.00276) and mixed distributions (MSE = 0.00366), lowering all three loss metrics (MSE, RMSE, MAE). For log norm data, subtracting the mean resulted in the lowest average losses (MSE = 0.0047).

\begin{figure*}[!htbp]
        \includegraphics[width=0.196\textwidth]{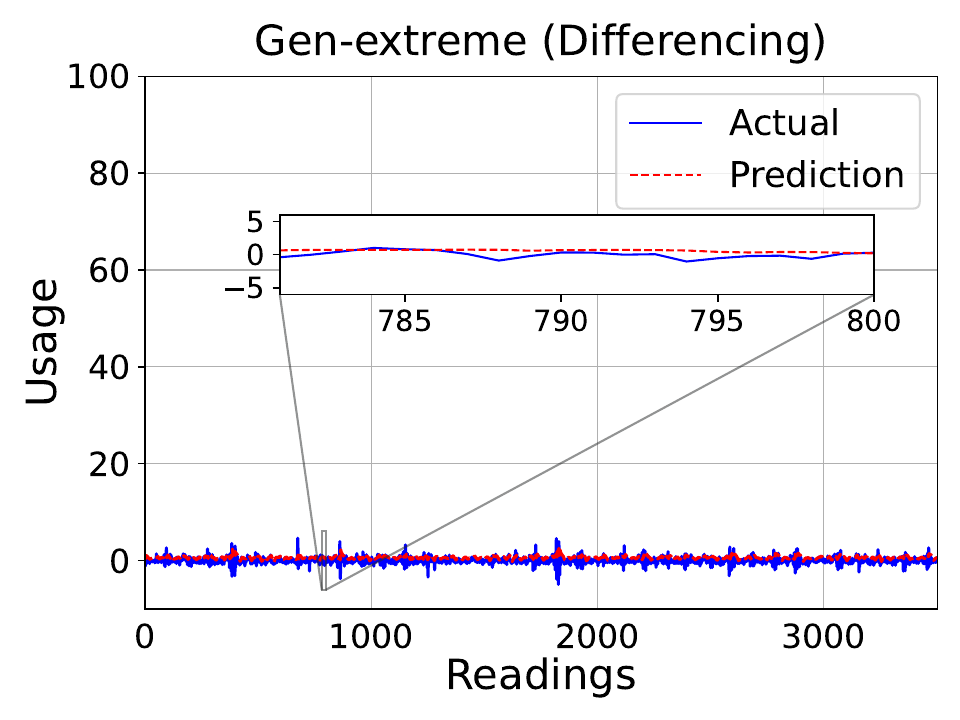}
        \includegraphics[width=0.196\textwidth]{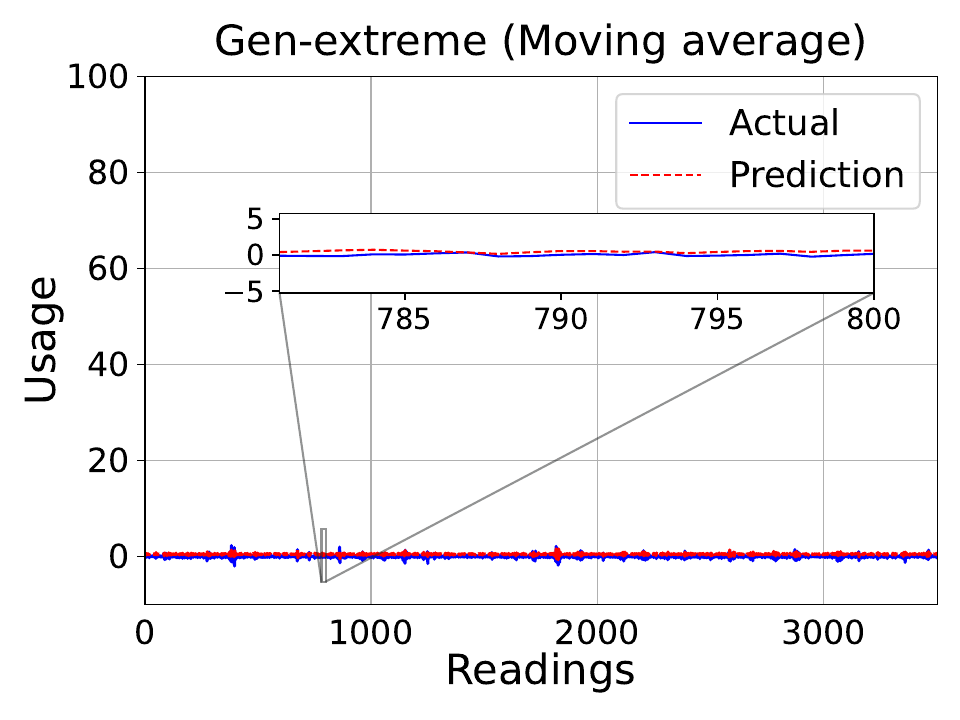}
        \includegraphics[width=0.196\textwidth]{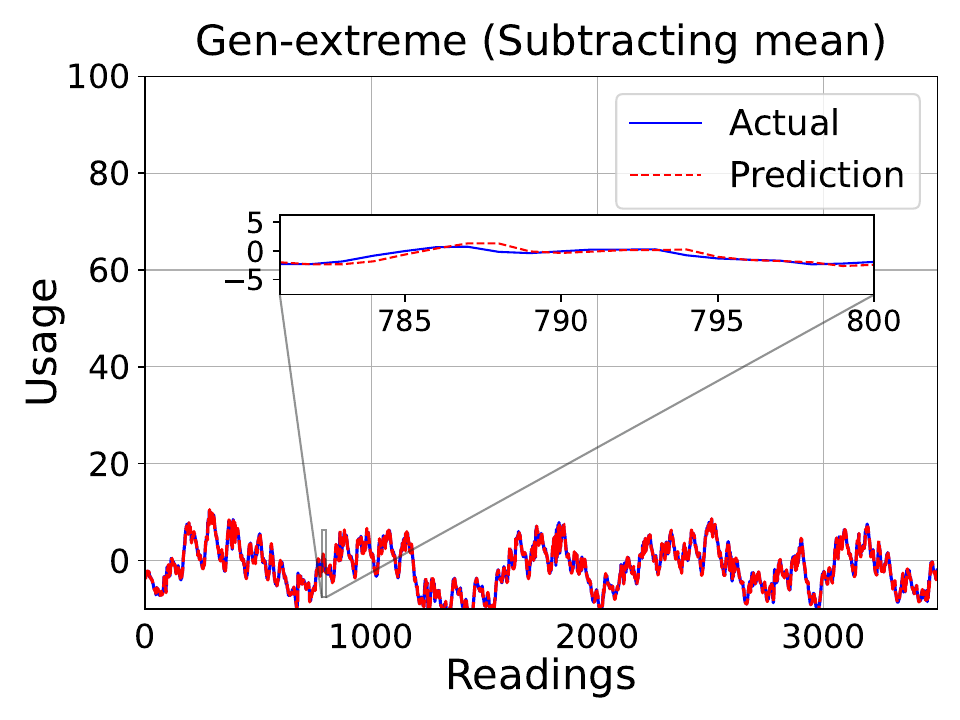}
        \includegraphics[width=0.196\textwidth]{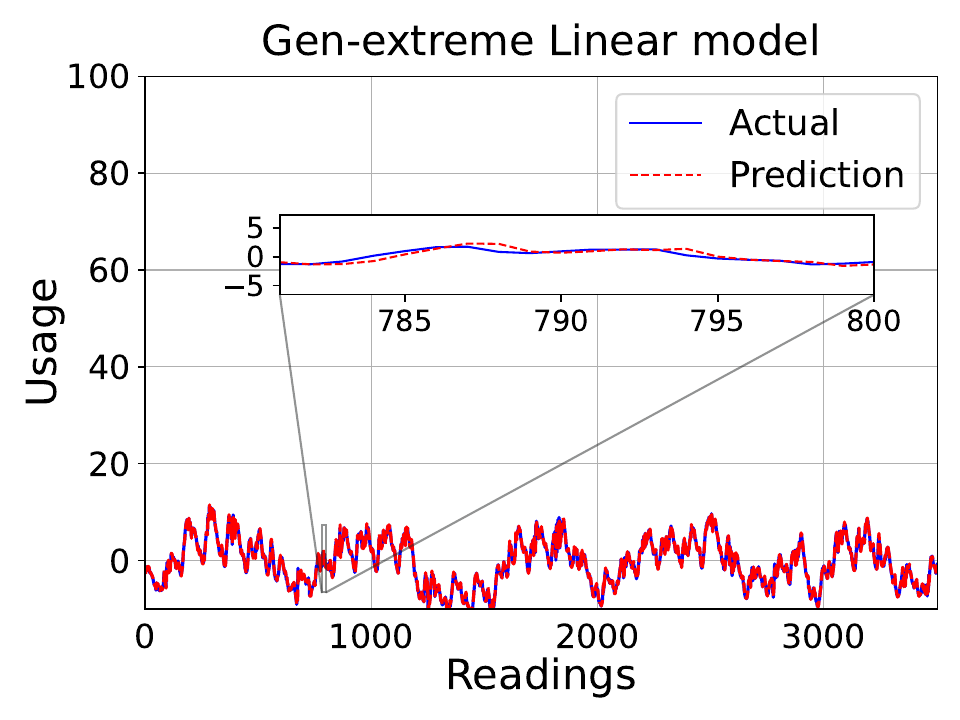}
        \includegraphics[width=0.196\textwidth]{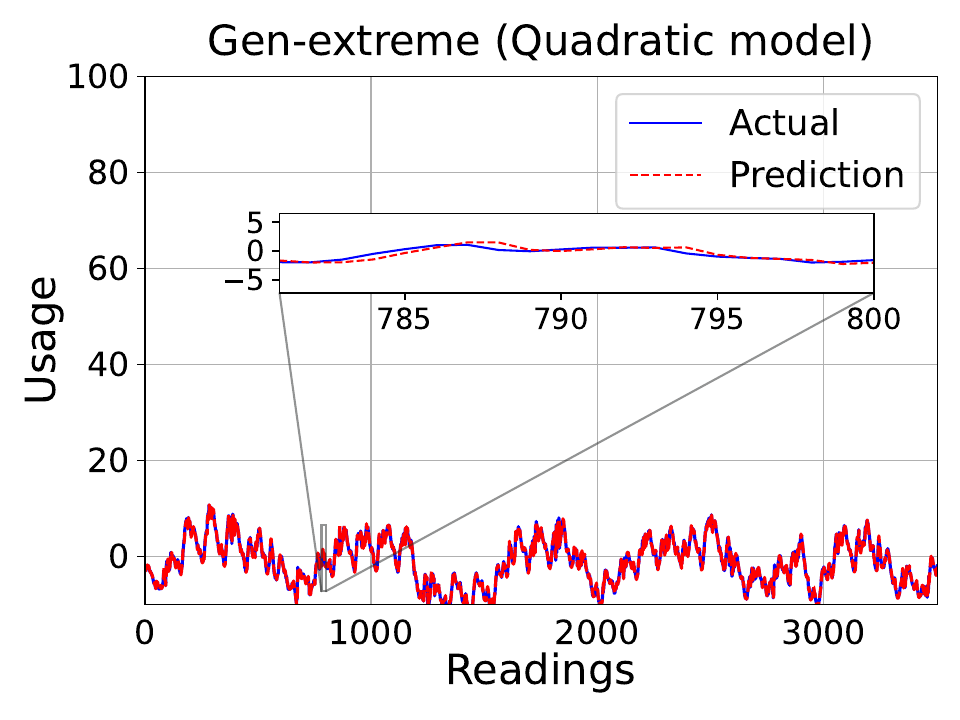}
    \caption{Model forecasting using FL approach for client 1, based on Ausgrid data following a gen-extreme distribution. The application of various detrending techniques is demonstrated, including differencing, moving average, mean removal, linear model, and quadratic model (from left to right).}
    \label{fig:ausgrid_c1_forecast}
\end{figure*}

\begin{figure*}[!htbp]
        \includegraphics[width=0.196\textwidth]{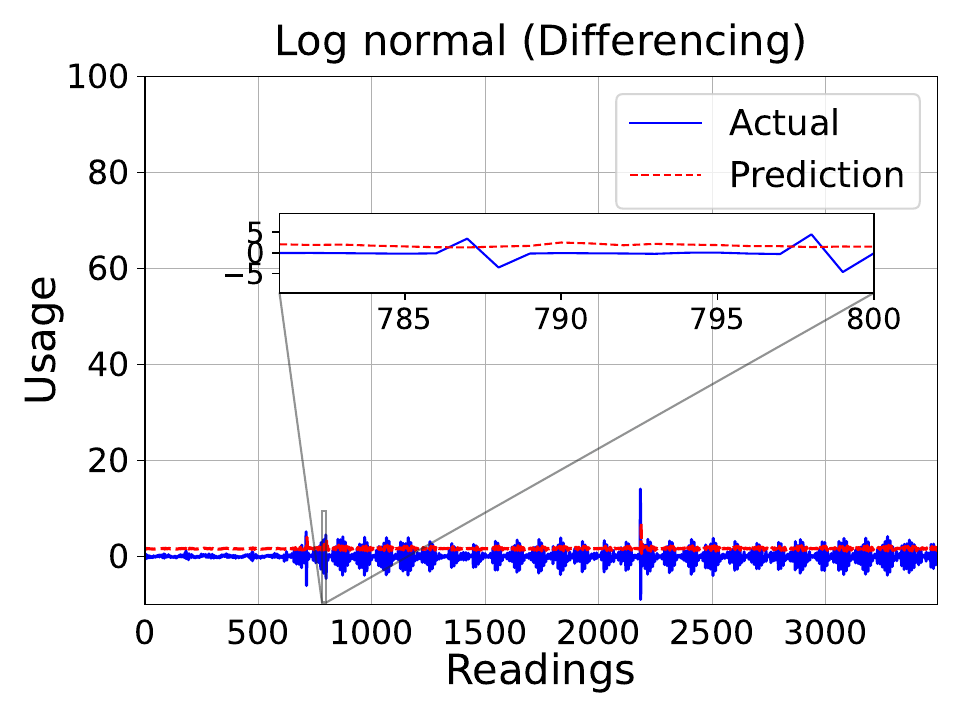}
        \includegraphics[width=0.196\textwidth]{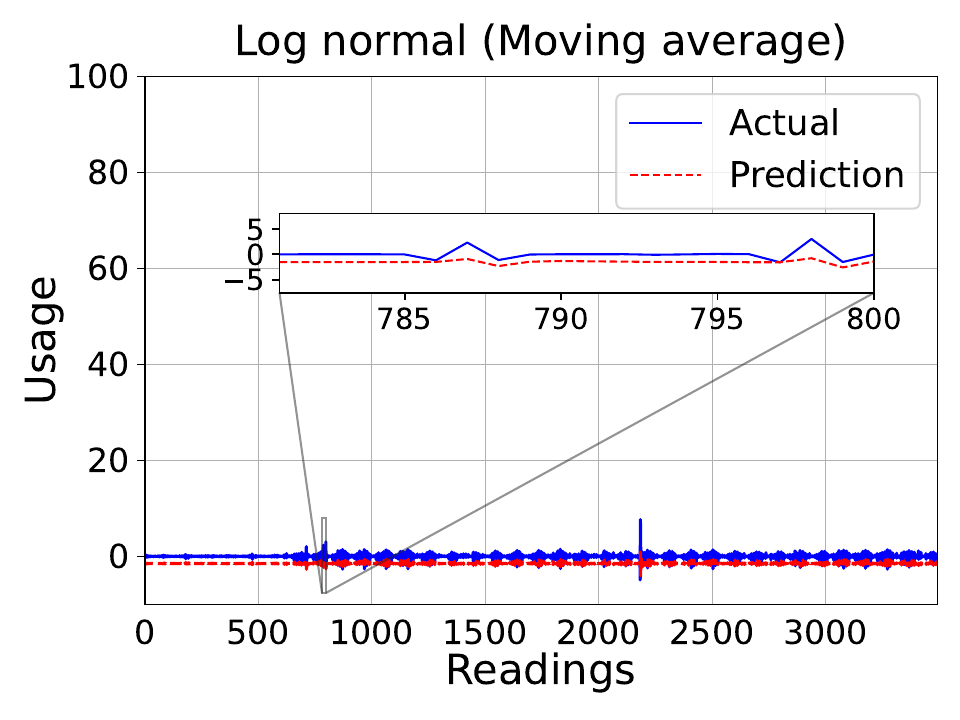}
        \includegraphics[width=0.196\textwidth]{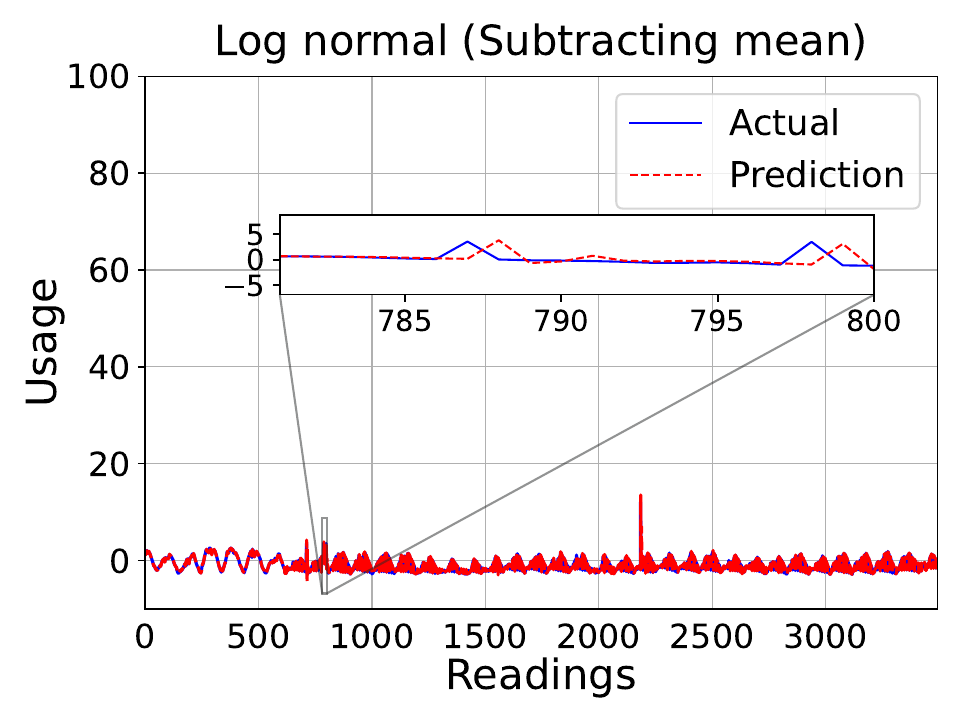}
        \includegraphics[width=0.196\textwidth]{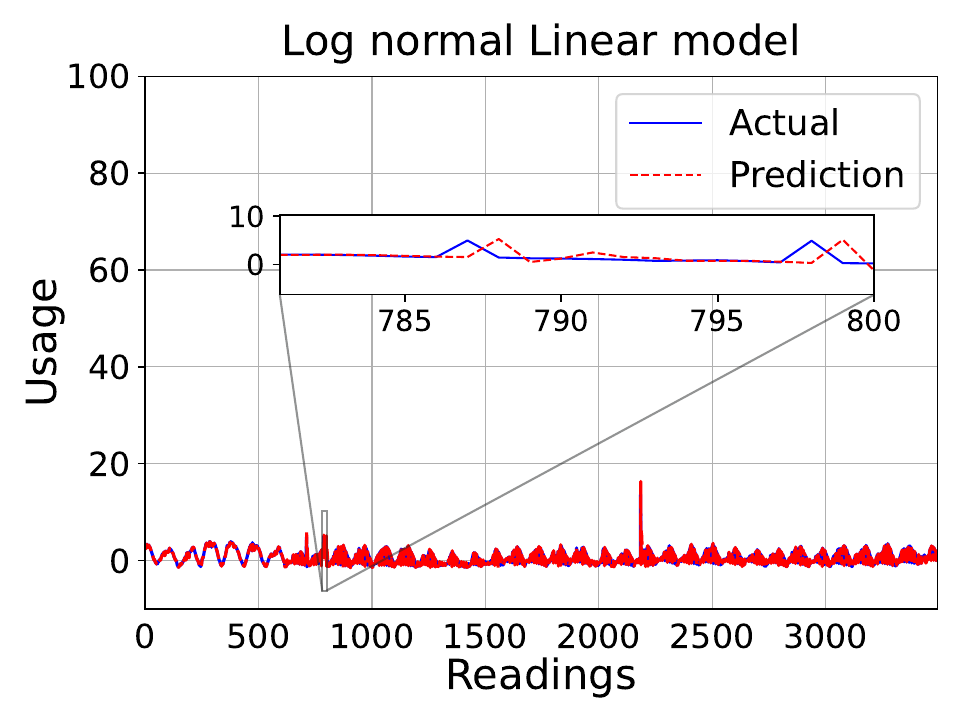}
        \includegraphics[width=0.196\textwidth]{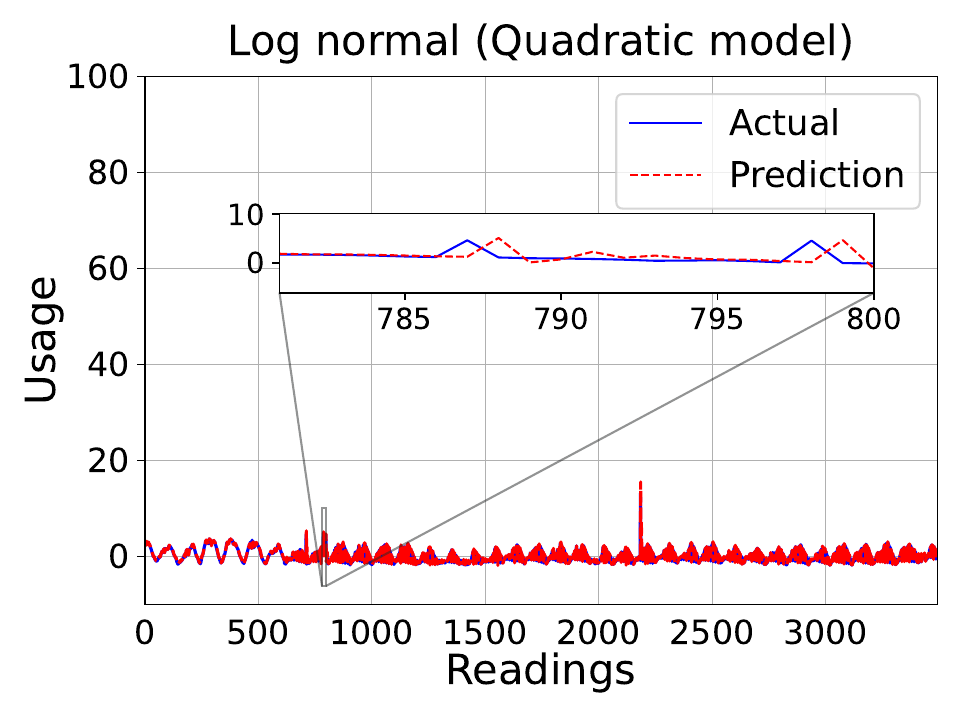}
    \caption{Model forecasting using FL approach for client 6, based on Ausgrid data following a log norm distribution. The application of various detrending techniques is demonstrated, including differencing, moving average, mean removal, linear model, and quadratic model (from left to right).}
    \label{fig:ausgrid_c6_forecast}
    \vspace{-10pt}
\end{figure*}

Overall, the results illustrated in Figures~\ref{fig:ausgrid_c1_forecast} and~\ref{fig:ausgrid_c6_forecast} highlight that selecting a suitable detrending method is critical. Inappropriate detrending can not only fail to improve performance but may also degrade forecasting accuracy.


\section{Conclusions}
\label{sec:conclusion}
A major contribution of this paper is to examine how a client's data distribution impacts the performance of an LSTM-based time-series forecasting model in both centralized and Federated Learning (FL) scenario. The presence of clients with non-linear data distributions, such as gen-extreme and log norm, significantly hampers FL performance in comparison to centralised settings. For instance, with a gen-extreme distribution, the MSE loss increased from 0.00273 to 0.00844, representing a 209.2\% rise. Similarly, for the log norm distribution, the MSE loss grew from 0.00589 to 0.01583, a 168.7\% increase. When comparing a gen-extreme distribution with a mixed distribution (gen-extreme and log norm) in FL, the loss surged from 0.00844 to 0.03115, marking a 269.1\% increase. Likewise, for the log norm distribution, the loss escalated from 0.01583 to 0.03115, an increase of 96.7\%.

Additionally, we underscore the significance of detrending for time-series data, a factor often overlooked in existing FL-based time-series forecasting studies. Our results show that applying an appropriate detrending technique led to a reduction in the average validation loss, thereby enhancing the overall forecasting performance. These findings were further validated with real-world datasets featuring non-linear distributions. For future work, we plan to investigate the characteristics of time-series data to develop data-driven client selection strategies and offloading approaches.


\bibliographystyle{unsrt}

\bibliography{IEEEabrv,ref}

\end{document}